\documentclass[sn-mathphys]{sn-jnl}
\usepackage{graphicx}
\usepackage{tablefootnote}
\usepackage{url}
\usepackage{natbib}
\usepackage{multirow}
\usepackage{array,booktabs}
\usepackage{pdflscape}
\usepackage{subfloat}
\usepackage{textcomp}
\usepackage{amsmath}
\usepackage{xcolor, soul}
\usepackage{xparse}

\sethlcolor{yellow}

\jyear{2021}%

\theoremstyle{thmstyleone}%
\theoremstyle{thmstyletwo}%

\theoremstyle{thmstylethree}%

\begin{document}

\title[Article Title]{An Exploratory Study on Utilising the Web of Linked Data for Product Data Mining}


\author*[1]{\fnm{Ziqi} \sur{Zhang}}\email{ziqi.zhang@sheffield.ac.uk}

\author[2]{\fnm{Xingyi} \sur{Song}}\email{x.song@sheffield.ac.uk}

\affil[1]{\orgdiv{Information School}, \orgname{University of Sheffield}, \orgaddress{\street{Regent Court, 211 Portobello}, \city{Sheffield}, \postcode{S1 4DP}, \state{South Yorkshire}, \country{UK}}}

\affil[2]{\orgdiv{Department of Computer Science}, \orgname{University of Sheffield}, \orgaddress{\street{Regent Court, 211 Portobello}, \city{Sheffield}, \postcode{S1 4DP}, \state{South Yorkshire}, \country{UK}}}


\abstract{The Linked Open Data practice has led to a significant growth of structured data on the Web. This has created an unprecedented opportunity for research in the field of Natural Language Processing. However, there is a lack of systematic studies on how such data can be used to support downstream NLP tasks. This work focuses on the e-commerce domain and explores how we can use such structured data to create language resources for product data mining tasks. To do so, we process billions of structured data points in the form of RDF n-quads, to create multi-million words of product-related corpora that are later used in three different ways for creating language resources: training word embedding models, continued pre-training of BERT-like language models, and training Machine Translation models that are used as a proxy to generate product-related keywords. These language resources are then evaluated in three downstream tasks, product classification, linking, and fake review detection using an extensive set of benchmarks. Our results show word embeddings to be the most reliable and consistent method to improve the accuracy on all tasks (with up to 6.9\% points in macro-average F1 on some datasets). The other two methods however, are not as consistent. Contrary to some earlier studies that suggest a rather simple but effective approach such as building domain-specific language models by pre-training using in-domain corpora, our work serves a lesson that adapting these methods to new domains may not be as easy as it seems. We further analyse our datasets and reflect on the literature to discuss potential issues that may help inform future research.}




\maketitle

\section{Introduction}
Recent years have seen significant increase in the adoption of the Linked Open Data practice \cite{bizer2009} by data publishers on the Web. LOD refers to the practice of describing structured data using standard markup languages (e.g., RDFa\footnote{\url{https://www.w3.org/TR/rdfa-primer/}}) and universal vocabularies (e.g., schema.org) that allow defining properties and relations of data, and publishing and interlinking such data on the open Web. While early LOD datasets primarily took form of a graph database such as DBpedia\footnote{\url{https://wiki.dbpedia.org/}}, an increasingly popular decentralised approach has been the embedding of semantic markup data within HTML pages. The Web Data Commons\footnote{\url{http://webdatacommons.org}} (WDC) project extracts such structured data from the CommonCrawl\footnote{\url{https://commoncrawl.org/}} as RDF n-quads\footnote{\url{https://www.w3.org/TR/n-quads/}}, and release them on an annual basis. As by September 2020, its statistics showed tehat almost 65\% of web pages, or 57\% of websites contained semantic markups, amounting to over over 98 billion quads - more than doubled that from 2019. 

The abundance of such structured data enables semantification of the Web in a way that can be read and understood automatically by computers. This has not only driven the development of new products and data integration services, but also created unprecedented opportunities for research in the areas of Natural Language Processing (NLP, e.g., \cite{foley2015,vagliano2017}). This is because the RDF n-quads extracted from such data are described by universal vocabularies that define concepts, their properties and relationships. One of the most popular vocabularies is schema.org, which currently contains nearly 800 concepts and 1,400 properties, and is used by over 10 million websites\footnote{\url{https://schema.org/}}. Therefore, an RDF n-quad can be considered as describing a `fact' that are instantiations of concepts, properties and relationships. Studies have shown that such data can be used to train models for various NLP tasks, such as event extraction \cite{foley2015} and entity linking \cite{bizer2019}.

A particular domain that is witnessing the boom of semantic markup data is e-commerce, where online shops are increasingly embedding structured product data described using schema.org vocabularies into their web pages in order to enable search engines to easily identify product offers and potentially drive traffic to their respective websites. Among the over 98 billion RDF n-quads mentioned above, nearly 10\% are related to products, and are described by schema.org vocabularies. As examples, previous studies \cite{Meusel2015,bizer2019} showed that among all product offers, 95\% had an n-quad related to their names, 65\% had one for their description, 35\% had one for their brand, and less than 10\% had one for their category. Such structured product data can be potentially useful as language resources to benefit various product data mining tasks. In particular, as the current Natural Language Processing research is shifting towards using large scale unstructured texts to build domain-specific language resources such as various domain-specific BERT models \mbox{\cite{Lee2019,alsentzer2019,beltagy2019,ZhangD2020,zhang2021}}, it would be interesting to know to what extent the previous success can be replicated in this domain, through exploiting this gigantic data. However, despite a limited number of sporadic studies \mbox{\cite{Meusel2015,bizer2019,zhang2020}}, there is a lack of work specifically targeting this direction.

In this work, we report our exploration of these problems in the product domain, which is chosen as the focus for two reasons. On the one hand, as discussed above, it is one of the most promising domains where a `critical mass' of such resources has been created. On the other hand, it is a domain that continues to garner interest from both researchers and practitioners, as evidenced by a series of workshops and shared tasks sponsored by industries \cite{Lin2018,zhang2020}. Specifically, we explore three different methods of using the semantic markup data of products to create different types of language resources: training word embedding models, continued pre-training of BERT-like language models, and training Machine Translation models that are used as a proxy to generate product-related keywords. We then evaluate such resources on three tasks related to product data mining: product classification, product linking, and fake product review detection. Surprisingly, we obtained overall negative results. Among the three methods, only word embeddings are the most reliable and consistent method to improve the accuracy on all three tasks, with a maximum of up to 6.9 percentage points in macro-average F1 on some datasets. The BERT language models and the MT-based product keywords on the other hand, do not bring improvements but can cause accuracy to decline. Although there have been many studies that successfully developed in-domain BERT models following the simple principle of continued pre-training a generic BERT using large domain-specific corpora, our results serve a lesson that this method may not be as easy as it seems. Moreover, there may be a need for further research to better understand the conditions for both the data and the process in order to ensure successful adaptation of such methods to new domains. To shed some light on these issues, we reflect on the previous literature and conduct a number of analyses of the data. We argue that the biased domain representation in the data and lack of vocabulary coverage could be attributing factors, and methods such as BERT language modelling and machine translation may be potentially more susceptible to such `data quality' issues than word embedding modelling. All of our data and resources are public and can be obtained upon request \footnote{\url{https://bit.ly/2MGpBR2}}. 



We organise the remainder of this paper as follows. Section \ref{relatedwork} reviews related work. Then Sections \ref{exp_prodcls} to \ref{exp_fakerev} presents our exploration on each of the three tasks. Each section will introduce our methodology, experiments, and presents the results. Section \ref{discussion} discusses the findings with further data analyses, and this is followed by a conclusion of our study in Section \ref{conclusion}. 

\section{Related Work} \label{relatedwork}
LOD resources have been around for decades and there have been a large number of studies \cite{Mountantonakis2020,Rama2020} and organised events\footnote{\url{http://events.linkeddata.org/ldow2018/}, \url{https://ld4ie.github.io/}} on the creation and consumption of such resources. However, previous studies have predominantly looked at LOD resources that are published as a graph database such as the DBpedia and Wikidata, while very few focused on the LOD published as semantic markup data embedded within web pages (to be referred to as `semantic markup data' in short). Although in both cases, data can be represented as generic RDF triples or n-quads, the fundamental difference is quality, which underpins the approaches that one can take to use such resources. Most of the LOD graph databases are well-curated, documented and maintained by organised efforts. Semantic markup data however, can be highly heterogeneous, noisy, and unbalanced \cite{kiran2020}, as they are created by significantly more decentralised efforts. There are also a blend of studies that focused on creating further LOD resources out of existing ones \cite{Chortaras2018}, compared to those that actually use LOD as language resources for downstream language processing tasks. Our literature review therefore has a specific focus on the following areas: 1) work that uses semantic markup data to create language resources, and 2) work on the three tasks we focus on, i.e, product classification, linking, and fake product review detection. While our work is also broadly related to the use of neural networks in creating domain-specific language models, such as the BioBERT \cite{Lee2019}, Clinical BERT \cite{alsentzer2019}, SciBERT\cite{beltagy2019}, E-Bert\cite{ZhangD2020}, and SMedBERT\cite{zhang2021}, these studies did not use semantic markup data and therefore, we do not expand our literature review to this broad area as that would significantly increase the scope of our discussion. However, in Section \ref{discussion}, we discuss our results with respect to the findings of these earlier studies. 

\subsection{Semantic markup data as language resources}\label{relatedwork_lr}
Research on using semantic markup data for downstream language processing tasks has just taken off in the recent years and therefore, studies addressing the creation of language resources from such data are limited. \cite{Primpeli2019} adopted an unsupervised approach to create a very large training dataset for product entity linking using semantic markup data extracted from the 2017 CommonCrawl corpus. The process started with extracting product offers that contain product identifiers annotated using the schema.org vocabulary. Then offers with the same identifiers are placed in the same cluster, followed by a cleaning process to eliminate potentially noisy clusters. The end clusters are considered to be product offers referring to the same product entity, and are used to train entity linking models. The work is further extended in later studies by \cite{bizer2019}, \cite{Peeters2020} and \cite{peeters2021}, and it was shown that this automatically created training dataset has a high quality and can be used to effectively train product entity matchers at high accuracy. While these studies investigated ad-hoc usages of semantic markup data as training data for specific tasks, our work explores possibilities of utilising such data to create language resources that are usable by a wider range of tasks.

The same authors also used a product corpus to train a domain-specific word embedding model in \cite{Peeters2020} using the fastText model. Specifically, they extracted the brand, name and description properties annotated by schema.org from the same corpus above, to create a text corpus that is used to train fastText embeddings. This domain specific embedding model gained minor improvement over a generic fastText embedding model on some product linking tasks. In comparison, this work also explores using product related corpora to train word embedding models. However, we study if this can generalise to other product data mining related tasks.

Work that uses semantic markup data to train embedding models can be traced back to \cite{Ristoski2018}, where authors used schema.org annotations (names and descriptions) of product entities to train entity embeddings using the paragraph2vec model \cite{Le2014}. This approach suffers similar limitations as the above, where the embeddings learned in such a way are ad-hoc and can only be used for entities found in the training process of the embedding models. Our study explores more generic ways of learning word embeddings.

In the 2020 Semantic Web Challenge on product data mining (MWPD2020, \cite{zhang2020}), a corpus of 1.9 billion words extracted from the descriptions of product entities annotated by schema.org vocabulary was used to train word embedding models. Compared to generic word embedding models, such models contributed to better results on the product classification task when used with a fastText baseline \cite{Lin2018}. However, they were not used by any of the participating teams of the shared task. This study fills this gap by thoroughly evaluating them on several product data mining tasks.

\subsection{Product classification}\label{relatedwork_pc}
Product classification is typically treated as an entity classification task, as the process involves assigning category labels (i.e., classes) to product instances based on their attributes - which we refer to as metadata - typically text-based such as name, description, and brands. Such labels usually reside in a categorisation taxonomy, therefore the task usually requires assigning multiple labels, one from each level of the taxonomy \cite{Lin2018}. Automated product classification is particularly useful for data integration services such as price comparison and integrated shopping platforms that need to organise product offers from heterogeneous sources into a universal catalogue. Below we summarise related work from different perspectives and highlight their similarities and differences.

\textbf{Metadata.} To classify products, features must be extracted from certain product metadata. Rich, structured metadata are often not available. Therefore, the majority of literature have only used product names, such as \cite{Kozareva2015,Chavaltada2017,Xia2017,Akritidis2018} and all of those participated in the 2018 Rakuten Data Challenge \cite{Lin2018}. Several studies used both names and product descriptions \cite{Cevahir2016,Gupta2016,Lee2018,Li2018,Borst2020,Li2020,Zehera2020}, while a few used other metadata such as model, brand, maker, etc., which need to be extracted from product specification web pages by an Information Extraction process \cite{Ha2016,Ristoski2018}. In addition, \cite{Ristoski2018} also used product images. The work by \cite{Meusel2015} and \cite{Zhang2019} used product categories allocated by the vendors and embedded as semantic markup data within the web pages. To differentiate these from the classification targets in such tasks, we refer to these as `site-specific product labels' or `categories'. The authors noted that despite the highly heterogeneous nature of such site-specific labels across different websites, they are still very useful for supervised classification. In comparison, this work explores a `new' type of metadata - product-related keywords generated by a machine translation model trained on the massive product corpora. Compared to product metadata already existing in a dataset and are comparatively better quality, such keywords may be very noisy. Our work will be first to explore if these keywords generated in such a way can be useful for product classification.

\textbf{Feature representation.} Generally speaking, for text-based metadata, there are three types of feature representation. The first is based on Bag-of-Words (BoW) or N-gram models, where texts are represented based on the presence of vocabulary in the dataset using either 1-hot encoding or some weighting scheme such as TF-IDF \cite{Cevahir2016,Ha2016,Chavaltada2017,Akritidis2018}. This often creates high-dimensional sparse feature vectors. The second uses pre-trained word embeddings or Language Models (LM) to create a relatively low-dimensional, dense feature vector of the input text. Word embeddings are vectorial representation of words pre-trained on very large corpora. Certain techniques will need to be applied in order to compose embeddings for long text passages based on single words. For example, \cite{Kozareva2015} averaged the embedding vectors of composing words from product titles, \cite{Lee2018} summed them, while work in \cite{Kim2014} (non-product domain) and \cite{Zhang2019} joined word embedding vectors to create a 2D tensor to represent the text. In the more recent work that uses pre-trained LMs such as BERT (e.g. in \cite{Yang2020}), the construction of text passage embeddings is taken care of dynamically by passing the input text through the language models directly, which will take into account the context of words. The third applies a separate learning process to learn a continuous distributional representation of the text directly from the downstream training datasets. This includes adaptation of the well-known Paragraph2Vec model such as in \cite{Gupta2016,Lee2018,Ristoski2018}. Our work will make use of feature representation methods from the second and the third types. However, studies by \mbox{\cite{Kozareva2015,Lee2018,Zhang2019,Yang2020}} used general-purpose, pre-trained word embeddings while we study the effects of embeddings purposefully trained on product related corpora. Studies by \mbox{\cite{Gupta2016,Lee2018,Ristoski2018}} created ad-hoc representations of product entities discovered in the training set, while such representations cannot be generalised to other data or tasks, our study explores more generic, data-agnostic methods for composing such representations.

\textbf{Algorithms.} The large majority of work has used supervised machine learning methods. These include those that use traditional machine learning algorithms \cite{Meusel2015,Cevahir2016,Chavaltada2017,Gupta2016,Lee2018,Ristoski2018}, and those that apply DNN-based algorithms \cite{Xia2017,Ha2016,Lin2018,Zhang2019}. All DNN-based methods have used some adaptations of CNN or RNN. These include the majority of the participating systems in the 2018 Rakuten Data Challenge. Besides, \cite{Meusel2015} also explored unsupervised methods based on the similarity between the feature representations of a product and target classes. \cite{Akritidis2018} studied product clustering, which does not label the resulting product groups. These represent unsupervised methods. Further, \cite{Li2018} studied the problem as a machine translation task, where the goal is to learn the mapping between a sequence of words from product names, to a sequence of product categories. 

MWPD2020 \cite{zhang2020} showed a trend towards using the pre-trained LMs for classification, such as those based on the BERT model \cite{devlin2018}. Such LMs are essentially deep neural networks trained by `reading' extremely large corpora containing billions of words, and are therefore, considered to have captured the `meaning' of human language to a certain extent. LM-based classification typically adds task-specific layers for the down-stream classification task on top of the pre-trained LM and then train the new model on downstream task data. Then typically, only the task-specific layers are learned from scratch. All participants but one at the product classification task at MWPD2020 used LM-based classification methods. The work by \cite{Yang2020} for example, used an ensemble model combining 17 different variants of the BERT model to achieve the best result on this task. 

This study does not focus on developing novel algorithms but instead, reuse existing ones such as the fastText baseline in \mbox{\cite{Lin2018}} and the DNN structure in \mbox{\cite{Zhang2019}}. It may however, reveal which algorithms are more susceptible to the different language resources created by this study.

\subsection{Product linking}\label{relatedwork_pl}
Product linking or matching is the task of determining if multiple product offers found from different websites (sometimes even from the same website) refer to the same, identical product entity. Since online shops will typically sell the same products, but publish different information about them, this task is particularly useful for services that need to collect and compare offers of the same products online. 

Product linking can be achieved by one of the three approaches: classification (e.g., \cite{zhang2020}), where product offer pairs are created a-priori and are classified to match or non-match; clustering, where a dataset of product offers are split into groups and members within the same group are considered to be about the same product; or retrieval (e.g., \cite{Li2020}) where the goal is to find the matching product entity from an existing database for a given product offer. In both classification and retrieval, often a `blocking' process is applied to reduce the search space. All three approaches depend on the calculation of `similarities' between product offers and this makes use of product metadata. A good literature review on product linking can be found in \cite{Ristoski2018}. Here, we summarise them in terms of metadata, feature representation, and algorithms in a similar fashion as before. 

\textbf{Metadata.} Similar to product classification, typically product linking will make use of product names (e.g., \cite{Kannan2011,Gopalakrishnan2012,Vandic2012,Bezu2015,Shah2018,Tracz2020,Li2020}) and descriptions (e.g., \cite{Petrovski2014,Ristoski2018,Li2020}). The difference however, is that the task also makes use of a diverse range of structured product attributes (e.g., \cite{Bezu2015,Shah2018,Petrovski2020,Li2020}), often defined as `key-value' pairs such as those that can be extracted from product specifications (e.g., product ID, model, brand, manufacturer). Intuitively, offers that have the similar sets of key-value pairs are more likely to match. Since such structured key-value attributes are often unavailable, many studies focused on how to extract them from the descriptions of an offer \cite{Kannan2011,Kopcke2012}, or from the specification table of the source web page \cite{Ristoski2018}. A small number of studies \cite{Kiapour2015,Wang2016,Ristoski2018} also used product images. Similar to product classification, we will explore the usefulness of product-related keywords generated by a machine translation model trained on the product semantic markup data. This has not been explored before.

\textbf{Feature representation.} Again, similar to product classification, broadly speaking, transforming textual metadata into feature representations is typically based on BoW (e.g., \cite{Vandic2012,Bezu2015}), pre-trained word embeddings or language models (e.g., \cite{Ristoski2018,Shah2018,Li2020,Peeters_intermediate_2020,peeters2021,Tracz2020}), or learning word embeddings on-the-spot from the downstream task datasets (e.g., \cite{Shah2018}). However, depending on the types of metadata, different methods may be adopted and then combined \cite{Kopcke2012}. For example, structured key-value attributes are often kept as-is and compared as a BoW, particularly if the values are short (e.g., product IDs). In \cite{Bezu2015}, a concept of `q-gram' was introduced to represent short texts (especially key-value pairs) as a set of character n-grams. Longer texts such as descriptions are better represented using word embeddings or through LMs. In this direction, similar sets of methods to product classification are used, e.g., aggregation of word embeddings, learning ad-hoc `product' representations, and dynamically constructed embeddings using LMs. For image data, typical pixel-based image representation approach is widely used \cite{Kiapour2015,Wang2016,Ristoski2018}. Same as product classification, in terms of novelty, our work focuses on evaluating the word embeddings purposefully trained on product related corpora while many earlier studies used generic word embeddings. We also use more generic methods for composing feature representations for product linking while previous models tried to learn `ad-hoc' representations.

\textbf{Algorithms.} Since the prediction of linking/matching of product offers depends on a notion of `similarity', some methods will have an `intermediary' step that converts product metadata features to similarity features \cite{Vandic2012,Li2020,Petrovski2020}. This is typically done by applying similarity metrics - usually based on string form, or word/character distribution - to the textual feature representations of two offers. Again, depending on the metadata, different similarity metrics may be applied \cite{Bezu2015,Kopcke2012,Li2020,Vandic2012}. This `intermediary' process creates a feature vector consisting of similarity scores computed by different measures, or using different features. The vector is then subject to another process to determine if the two offers should match. However, as mentioned before, some methods \cite{Shah2018,Petrovski2020} do not require such an intermediary step, as the similarity computation is embedded as part of the method that tackles the task in an `end-to-end' fashion.

In terms of the method to the end-task, most studies are based on supervised binary classification, which aims to determine if a pair of offers should match or not. Following a similar pattern to product classification, the classification algorithms have evolved from the traditional \cite{Kannan2011,Petrovski2014,Li2020}, to DNN-based \cite{Shah2018,Li2020}, to LM-based \cite{Borst2020,Peeters_intermediate_2020,peeters2021,Tracz2020}. Depending on the classification algorithm, the input could be the similarity feature vector of a pair (e.g., \cite{Kannan2011}) computed by the intermediary step, or directly the feature vectors derived from the metadata of each offer \cite{Shah2018,Peeters_intermediate_2020,peeters2021}. In the second case, DNN- or LM-based algorithms are typically used, where the network will consists of two channels, each learning feature representations from one offer and they are then combined into a single channel feeding into the classification layers. In the study by \mbox{\cite{peeters2021}} which extends their earlier work in \mbox{\cite{Peeters_intermediate_2020}}, the authors proposed a multi-task learning neural network based on the BERT model, tailored for the product linking task. In addition to learning to predict if two pairs of product offers refer to the same entity (binary classification), the model at the same times learns to predict the shared product identifier by the two offers (multi-classification). Additionally, one can also make use of cut-off thresholds of similarity to determine match/non-match \mbox{\cite{Gopalakrishnan2012}}.

Clustering is used in a number of studies, such as \cite{Bezu2015} that clustered offers based on the `q-grams' derived from their names and key-value pairs; and \cite{Londhe2014} where a `strength of ties' style of clustering was applied to a `network' of important words derived from a pair of product offers to determine if they form a cohesive `community' and therefore, should match. 

Methods that require offer-pairs as input will often require a `blocking' pre-process. This is because in practice, we often face the task of discovering matching offers from a collection of offers, where the potential matching pairs are unknown and the computation could become intractable if we were to analyse all pair-wise combinations. The blocking process aims to reduce the search space for pairs, to create a minimal set of pairs for classification. Blocking strategies are varied and often lightweight, such as \cite{Kopcke2012} that is based on matching manufacturers and categories, and \cite{Gopalakrishnan2012} that is based on string prefix.

A unique direction of research in product linking looks into automated expansion of training data, either in terms of training instances, or metadata that can be used for feature extraction. For example, \cite{Gopalakrishnan2012} enriched the name of product offers with tokens retrieved using a web search engine. \cite{Borst2020} used product offer names to fetch similar entities from Wikidata, to create additional training instances. 

Compared to state-of-the-art, we focus on the sub-task of supervised, binary classification of match/non-match, while ignoring the `blocking' process. Our method will use state-of-the-art algorithms, as our research focus is on evaluating the impact of the language resources created from the semantic markup data on existing algorithms.

\subsection{Fake product review detection}\label{relatedwork_fakeprodrev}
Fake reviews, as per \cite{WU2020}, generally refer to reviews created in an attempt to mislead consumers (either in a positive or negative way). They are also known as deceptive opinions, spam opinions, or spam reviews \cite{Mandhula2020}. Fake online reviews in e-commerce significantly affect consumers, merchants, and market dynamics. In extreme cases, they led to financial loss for companies and legal cases \cite{Gani2015}. While traditionally, fake reviewers are written by humans, with the advancement of Natural Language Generation technology, it has been shown that fake reviews automatically generated by programs are even more difficult for human annotators to detect \cite{Salminen2022}. There is an extensive amount of studies on automated fake review detection and for that reason, we refer readers to the surveys by \cite{WU2020} while below we present a brief overview of this field, highlighting the novelty of our work. Further, in addition to studies focusing on detecting the content, there are work (e.g., \cite{Xu2019,Zhang2018}) that detect spammers (users) and spammer groups (network) which we do not cover here. 

Detecting fake review is predominantly treated as a supervised, binary text classification task. Thus similar to product classification, it involves extracting features of the review text (metadata), representing them in a machine processable format (feature representation), and training a model that is able to generalise patterns using the features and apply the patterns to unseen data (algorithm). In terms of \textbf{features (metadata)}, \cite{Salminen2022} broadly categorised them into `lexical' and `non-lexical'. Lexical features are attributes derived from text, such as words, n-grams, punctuations and latent topics. Non-lexical features are metadata related to the reviews (e.g., ratings, stars) or their authors (ID, location, number of reviews generated). In terms of \textbf{feature representation} and \textbf{algorithms}, same patterns to that of product classification are noticed due to the two tasks been handled by text classification approaches. Briefly, research has evolved from early methods that use manually engineered features in a 1-hot encoding (e.g., \cite{Ball2014}) to pre-trained word embeddings (e.g., \cite{Ren2016,Yuan2018}), and learning representations of the target dataset as part of the model, on the spot (e.g., \cite{Liu2019b}). The use of machine learning algorithms also evolved from the earlier classic algorithms such as SVM and logistic regression (e.g., \cite{Ball2014}), to deep neural networks (e.g., \cite{Ren2016,Yuan2018}), to using very large LMs such as BERT (e.g., \cite{Salminen2022}). 

Compared to the previous studies, our work does not aim to introduce new features or algorithms. Instead, we explore the usefulness of the feature representations learned from massive product related semantic markup data on the task of fake review detection. Earlier work such as \cite{Liu2019,Yuan2018} used word embeddings pre-trained on general purpose corpora, and \cite{Ren2016} trained domain-specific word embeddings using an Amazon product review corpus. In contrast, our work is the first that explores if a corpus of product details (instead of their reviews) can be used to learn word embeddings for this task. Compared to \cite{Salminen2022} who also used LMs, our work explores the effect of continued pre-training of LM using in-domain corpus, while \cite{Salminen2022} did not.

\subsection{Reflection}\label{relatedwork_ref}
Summarising related work above, our study addresses two limitations of state-of-the-art. First, despite the abundance of semantic markup data on the Web, there are only a very small number of studies that explored the use of such data to create language resources for downstream language processing tasks. Among them, the typical approach is training embedding models using such data \cite{Ristoski2018,Peeters_intermediate_2020,zhang2020}. However, these methods and/or resources are often ad-hoc, and their effects have not been compared on the same tasks. 

Second, despite the continued interest in the research of product classification, linking, and fake review detection, the use of language resources to support such tasks has been highly inconsistent, ranging from no-use at all to using a diverse set of word embedding models (e.g., \cite{Kozareva2015,Zhang2019,Borst2020}). It is unclear for example, if earlier success of building domain-specific LMs by continued pre-training of BERT models using in-domain corpora can be replicated in this domain. Adding to this complexity is the use of different datasets, and diverse use of machine learning models ranging from traditional algorithms (e.g., SVM, logistic regression), to deep neural networks, to pre-trained language models. The implication of this is that it is extremely difficult to compare the effect of using certain language resources on such tasks.

Motivated by these issues, our work in the following will explore three different ways of creating language resource using semantic markup data, and systematically evaluate them under uniform settings on the three different downstream tasks mentioned above.

\section{Building Language Resources} \label{method_lr}
In this section, we describe our method for the creation and evaluation of the language resources for product data mining. We begin with introducing the data sources we use to create the language resources (Section \ref{method_data}). We then discuss three different ways of using these data sources to create different types of language resources: training word embedding models, continued pre-training of BERT-like LMs, and training Machine Translation models that are used as a proxy to generate product-related keywords. These language resources will be later used in the three downstream tasks, to be detailed in Sections \ref{exp_prodcls}, \ref{exp_prodmatch} and \ref{exp_fakerev}.

\subsection{Data sources}\label{method_data}
In order to create language resources using semantic markup data for the product domain, we used the 2017 release of the structured data crawled by the WDC project. Specifically, we only downloaded and processed the class-specific subsets of the schema.org data related to \texttt{sg:Product}\footnote{\url{http://webdatacommons.org/structureddata/2018-12/stats/schema\_org\_subsets.html}}. This contains nearly 5 billion RDF n-quads, extracted from over 267 million web pages and over 812 thousand hosts. Each n-quad contains a subject, predicate, object, and a graph-label which in this case, denotes the source URL of the n-quad. 

Next, we parse this dataset to identify product offer instances, and build a Solr\footnote{\url{https://lucene.apache.org/solr}} index of product offers with their attributes found in the n-quads. This is done by firstly searching for `definition n-quads' that define a product offer instance with \texttt{http://www.w3.org/1999/02/22-rdf-syntax-ns\#type} as the predicate, and either \texttt{sg:Product} or \texttt{sg:Offer} as the object (i.e., where an n-quad defines an instance of an \texttt{sg:Product} or \texttt{sg:Offer}), and then parsing other n-quads containing the same subject as the definition n-quad to create property-value pairs for each offer. Only data that are potentially English are retained. This is achieved by automatically checking if the source URL (i.e., graph label) contains a top-level domain that clearly indicates non-English websites (e.g., .fr, .cn). This Solr index is further processed to create two corpora: a product description corpus, and a product category corpus. 

The \textit{product description corpus} contains descriptions of product offers. These are extracted from the \texttt{sg:Product/description} property of each product offer. A light cleaning process is applied to ensure that only descriptions containing between 50 and 250 words are selected. This restriction is to reduce content that is likely to be very noisy. For example, we noticed that sometimes product descriptions contain only a handful of generic words; while other times they are too long and can include the entire web page content. These texts are also normalised to keep only alpha-numeric characters. If a token contains digits only, it is replaced with a symbolic token to indicate a digit-only token. The resulting product description corpus contains over 1.9 billion tokens, extracted from over 34 million product offers. 

The \textit{product category corpus} contains over 700 thousands of product name - site-specific category pairs. These are selected from offer instances that have both an n-quad defining their name and site-specific labels. Product names are extracted from \texttt{sg:Product/name} or \texttt{sg:Offer/name} properties, while site-specific labels are extracted from \texttt{sg:Product/category} or \texttt{sg:Offer/category}. Both product names and site-specific labels are subject to a light cleaning process where only alpha-numeric characters are retained, and those containing more than 10 tokens (delimited by white space characters) or less than 2 tokens are removed. These restrictions are for the same reason - to reduce noise in the data. Also, digit-only tokens are replaced with the same universal symbol. Further, a stop word list is used to filter out generic site-specific labels, such as \texttt{Home} and \texttt{Product}, and only pairs extracted from the top 100 largest hosts (as measured by the number of product offer instances found from each host) are kept. This is to focus on hosts that are potentially large e-commerce vendors and therefore, have defined relatively good quality site-specific categorisation schemata. 

We will explain how we use these corpora to build language resources below. 

\subsection{Training word embedding models}\label{method_lr_we}
The first approach to utilising the above corpora is training word embedding models. As discussed before, only a couple of studies \cite{Ristoski2018,zhang2020} used semantic markup data to train embedding models. However, \cite{Ristoski2018} trained product embeddings that are ad-hoc, while our earlier work \cite{zhang2020} developed word embedding models that were not thoroughly evaluated. Here, following our previous work, we simply use the Gensim\footnote{\url{https://radimrehurek.com/gensim/}} implementation of the Word2Vec algorithm \cite{Mikolov2013} to train word embedding models using the product description corpus. We use the skip-gram algorithm for training, as it was shown to better represent infrequent words \cite{Mikolov2013}. This fits our data well, as a notable fraction of words in our product classification and linking datasets (see Appendix \ref{appA}) are not represented by the most frequent words found in the product description corpus. 

We use a sliding window of 10, minimum frequency threshold of 5 and text lower-casing, then keeping the remaining parameters as default. The word embeddings have 300 dimensions. We refer to this as `product word embeddings'.

\subsection{Continued Pre-training of BERT language models}\label{method_lr_lm}
The second approach explores the continued pre-training of large LMs. The principle of `continued pre-training' of LMs has been introduced in the recent research. The idea is to take an existing LM such as BERT, and further training it on large, in-domain, unlabelled corpora (e.g., \cite{beltagy2019scibert,lee2020biobert}). 

We explore the benefits of continued pre-training the BERT model on our product description corpus, and refer to the resulting LM as `BERT\textsubscript{prod}'. Specifically, we take the \texttt{`bert-base-uncased'} model\footnote{\url{https://huggingface.co/bert-base-uncased}} and run the masked language modelling task on our product description corpus, keeping all hyper-parameters as the default\footnote{Using the implementation at \url{https://github.com/huggingface/transformers/tree/master/examples/language-modeling}}.

However, pre-training LMs is an extremely resource-demanding process, and due to our limited access to HPC resources, we had to split our product description corpus to small segments, and create different versions of BERT\textsubscript{prod} models. Specifically, we randomly sampled 8\% (or approx. 570MB, which is the maximum size of a corpus we can fit with the pre-training process on our hardware) of our corpus for 7 times ensuring no overlap of selected product descriptions, thus creating 7 smaller corpora to continue to pre-train the BERT model. This creates 7 BERT\textsubscript{prod} models, 
and the total size of data used for continued-pretraining represents 50\% of the original product description corpus. 

\subsection{Training machine translation models}\label{method_lr_mt}
The third approach to utilising the product corpora is inspired by the work of \cite{Li2018}. The authors cast product classification as an MT task, whose goal is to learn the mapping between the sequence of words in a product name, to the sequence of category labels that form a hierarchical path. In this sense, the product names and their category label paths are treated as two different languages. 

However, an important difference of Li's work from ours is that the dataset they used for training the MT models is arguably, much better quality. This is because it is collected from a single vendor website, hence there is only one categorisation scheme and the naming and categorisation of products are generally consistent. In contrary, our product category corpus contains data from hundreds of different hosts, potentially selling very different products, and therefore used highly different and inconsistent categorisation schemata, which will have different levels of hierarchies. Further, our goal of product classification is to assign category labels from a universal schema to products from different vendors. Therefore, the site-specific categories cannot be directly used as classification targets. 

Therefore, instead of using this corpus to directly train a product classifier, we use it to train MT models that map a sequence of words in the product name, to the sequence of words in the product's site-specific category. Then given a product name in the downstream task data, we apply the MT model to generate a sequence of words, which although will unlikely to map to the end classification labels, may still be indicative of the product's `type' or 'category' and therefore, become useful features for the downstream tasks. We will refer to these words as `product related keywords' (denoted as `pk'). 

To train the MT model, we apply off-the-shelf MT toolkit OpenNMT \cite{klein2018opennmt} to the product category corpus. The encoder and decoder are 2-layer LSTM with 500 hidden units. We use all default settings of of the hyperparameters in the distributed implementation.

\section{Product Classification}\label{exp_prodcls}
In this section, we explore the usage of the different language resources created in Section \ref{method_lr} in the task of product classification. We show the datasets used for this study, and configure a number of models and compare them to evaluate the impact of these language resources on these datasets. We then present the results, which will be further discussed later in Section \ref{discussion} together with results from other tasks. 

\subsection{Datasets}\label{exp_prodcls_data}
\begin{table}[thb]
	\centering
	\renewcommand{\arraystretch}{1.5}
	
	\begin{tabular}{>{}m{2cm} >{}m{1.0cm} >{}m{1.3cm} >{\arraybackslash}m{1.0cm}>{\arraybackslash}m{1.5cm}>{\arraybackslash}m{2.5cm}}
		\toprule    		    		    		
		Dataset    &Train    &Validation    &Test    &Classes    &  Metadata  \\
		\hline 
		Rakuten &800k	&n/a	&200k	&3,008 &name (\textbf{n})\\
		IceCat &489,902	&122,476	&153,095	&370 &name, description (\textbf{d}), brand (\textbf{b})\\
		WDC-25 &20,205	&n/a	&4,884	&25 &name, description, brand, manufacturer (\textbf{m})\\
		MWPD-PC &10,012	&3,000	&3,107	&lvl1=37, lvl2=76, lvl3=281 &name, description, site-specific label/ category (\textbf{c})\\
		\hline                          
	\end{tabular}
	\caption{Summary of datsets for product classification.}
	\label{tab_pc_dataset}
\end{table}

We use four datasets listed in Table \ref{tab_pc_dataset} the \textbf{Rakuten} dataset is the one used in the Rakuten Data Challenge (\cite{Lin2018}). This contains one million product offers crawled from Rakuten.com, an online e-commerce marketplace. Product offers are classified into a tree-based hierarchical taxonomy, however due to data confidentiality, all categories are pseudonymised, resulting in over 3,000 different category labels. The dataset is split into a training set of 800k, and test set of 200k. Each product offer only has one type of metadata, i.e., its name. 

The \textbf{IceCat} dataset is released under the WDC project\footnote{\url{http://data.dws.informatik.uni-mannheim.de/largescaleproductcorpus/categorization/}}, and contains over 760k product offers crawled from IceCat.de, a worldwide publisher and syndicator of multilingual, standardised product data from various domains. Product offers are classified into IceCat's categorisation system with over 300 different labels. The dataset is split into a training set of around 489k offers, a validation set of around 122k offers, and a test set of around 153k offers. Each offer has three types of metadata: name, description and brand.

The \textbf{WDC-25} dataset is also released by the WDC project\footnote{\url{http://webdatacommons.org/categorization/index.html}}, and contains around 24k product offers randomly sampled from over 79k websites. These are classified into a flat categorisation scheme of 25 different labels, developed with reference to the Amazon, Google and UNSPSC\footnote{\url{https://www.unspsc.org/}} product catalogue taxonomies. In the original work by WDC, this dataset was further compressed into over 2,900 product instances by identifying clusters of offers that refer to the same product instance and merging their metadata. In our work, we do not use this `compressed' dataset but the original dataset of 24k product offers as-is. This is split into a training set of over 20k offers and a test set of around 5,000 offers. Each offer has a large number of metadata but only the following are selected for this work: name, description, brand, and manufacturer.

The \textbf{MWPD-PC} dataset is the product classification dataset released in the MWPD2020 challenge (\cite{zhang2020}). It contains around 16k product offers randomly sampled from the structured product data (described by the schema.org vocabulary) crawled by the WDC project. These are classified into three levels of classification (lvl1 to lvl3) following the GS1 Global Product Classification standard (GPC) \footnote{\url{https://www.gs1.org/standards/gpc}}. The dataset is split into a training set of over 10k offers, a validation and test set each containing around 3,000 offers. Each offer has the following metadata: name, description, and site-specific label.

The product metadata are of various word lengths from different datasets. However, neural network based classification models require text input to be of a fixed length. The normal practice is that if a real input text is shorter than this fixed length, it is padded with `arbitrary' tokens. If it is longer, it is truncated. We configure the lengths according to Table \ref{tab_input_length_prodcls}, based on the longest input observed on the datasets and the corresponding metadata used. All the training, validation, and test splits are based on the original data releases. Our selection of datasets represents a significant degree of diversity. The Rakuten and IceCat datasets are created on single e-commerce website and therefore, may be considered `higher quality' as the offers may be defined more consistently. They are also much larger datasets. The WDC-25 and the MWPD-PC datasets contain data collected from a very large number of different websites and therefore, may be considered `noisier' due to the high level of heterogeneity in the dataset. Table \ref{tab_pc_dataset} shows the statistics of these datasets.

\begin{table}[thb]
	\centering
	\renewcommand{\arraystretch}{1.5}
	
	\begin{tabular}{>{}m{3.2cm} >{}m{6cm}}
		\toprule    		    		    		
		Dataset    & Product metadata and fixed length (tokens)            \\
		\hline 
		Rakuten &name (32)  \\
		IceCat & all (256); name (32)\\
		WDC-25 &all (256); name (32) 	  \\
		MWPD-PC &all (256); name (32) \\
		\hline                          
		
	\end{tabular}
	\caption{Configuration of input word length for neural network based classification models. `all' refers to concatenating all product metadata detailed in Table \ref{tab_pc_dataset} as a single text input.  }
	\label{tab_input_length_prodcls}
\end{table}

\subsection{Model configurations}\label{exp_prodcls_models}
\begin{figure}
	\centering
	\includegraphics[width=350pt]{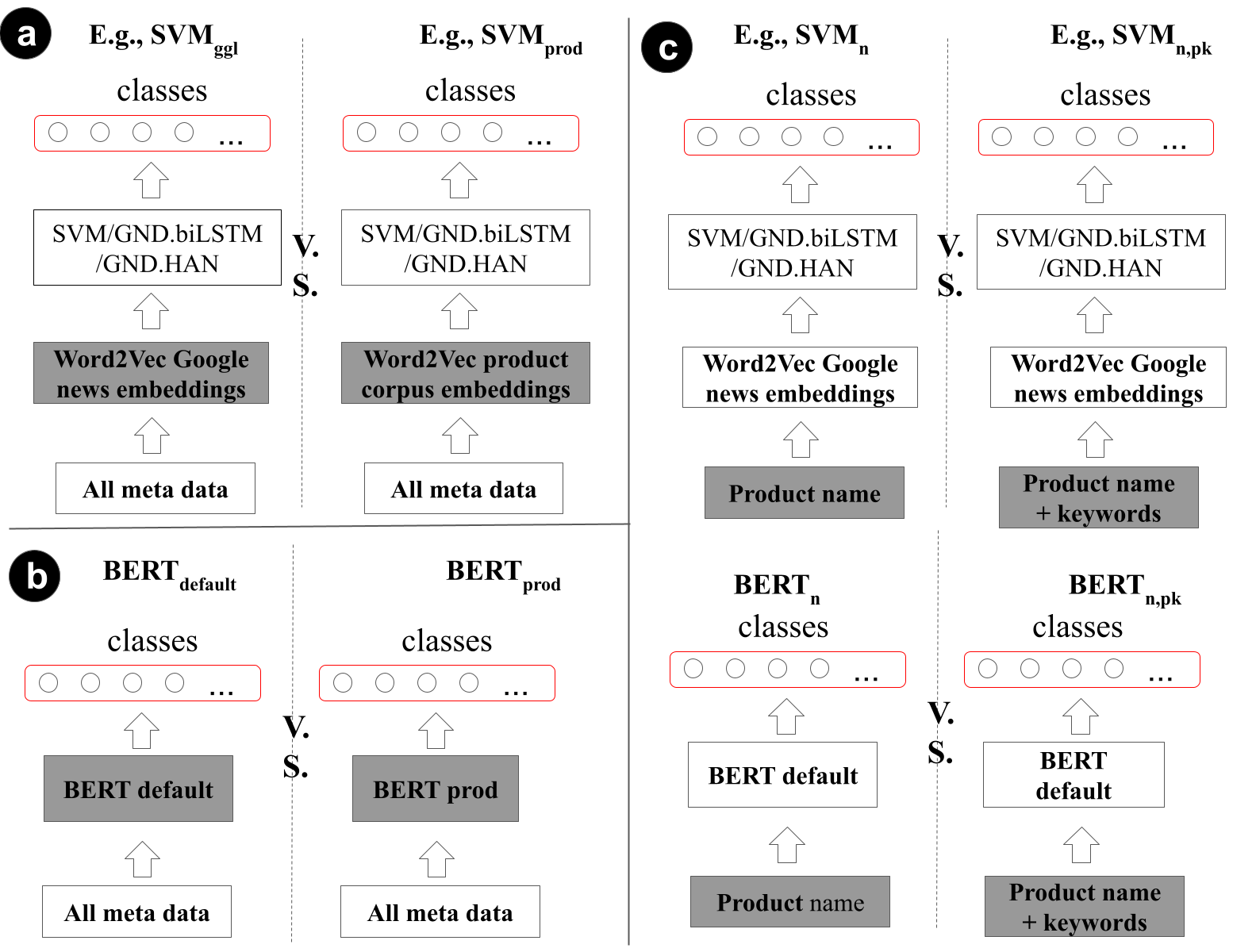}
	\caption{Configurations of different models for comparison for the product classification task. The shaded box represents the components of a model to be changed for comparison.}    
	\label{fig_prodcls_structure} 
\end{figure}

Models are configured based on the variations of the input product metadata, feature representation methods, and the machine learning algorithms. Figure \ref{fig_prodcls_structure} lists these models that will be discussed in detail below. 

\subsubsection{Using word embeddings}\label{exp_prodcls_models_wordemb}
Shown in Figure \ref{fig_prodcls_structure}a (Part (a) Experiments), the baseline and their corresponding comparative models differ in terms of word embedding representations (shaded in grey). Given a product, each model takes input all of its metadata available in a dataset, and passes them through the different word embeddings to construct a feature representation for the product. An ML algorithm then learns to classify the products based on these features. 

In terms of the word embeddings that are key for comparison, we compare our product word embeddings (\textbf{prod}) against the generic, pre-trained Word2Vec embedding model trained using Google News (\textbf{ggl})\footnote{\url{https://code.google.com/archive/p/word2vec/}}. In terms of ML algorithms, we test a simple linear SVM (\textbf{SVM}), the fastText baseline used in the MWPD2020 shared task for product classification \cite{zhang2020} (\textbf{FT.MWPD}), and the `GN-DeepCN' structure proposed in \cite{Zhang2019} with either a bi-LSTM (\textbf{GND.biLSTM}) or HAN (\textbf{GND.HAN}) as its sub-structure. For SVM and fastText, different product metadata are concatenated into a single text and treated indifferently. For GND.biLSTM and GND.HAN, each specific type of product metadata is fed into a sub-structure (biLSTM or HAN) that learns separate feature representations for them. The implementation and specifications of these algorithms are as follows:

\begin{itemize}
	\item SVM: implemented in Scikit-Learn 0.19, with the parameters set as follows: regularisation (\textit{c}) parameter of 0.01, one-vs-rest multi-classification training, balanced class weight, L2 penalisation and squared hing loss.
	\item fastText: default implementation as in \cite{zhang2020} 
	\item GND.biLSTM and GND.HAN: default implementation by \cite{Zhang2019}, using an epoch of 20 and a batch size of 128. All other hyperparameters remain unchanged. 
\end{itemize}

Following this, a model using the product word embeddings will be compared against itself when using the generic word embeddings. E.g., SVM\textsubscript{prod} against SVM\textsubscript{ggl}, or GND.HAN\textsubscript{prod} against GND.HAN\textsubscript{ggl}.


\subsubsection{Using language models}\label{exp_prodcls_models_lm}
Shown in Figure \ref{fig_prodcls_structure}b (Part (b) Experiments), the baseline and their corresponding comparative models differ in terms of the underlying language models (LM) used. Given a product, each LM takes input all of its metadata available in a dataset, and passes them into the different LM (shaded in grey), which produces its feature representation and learns classification patterns in an end-to-end fashion. 

In terms of the LM, we compare a generic BERT model (\textbf{BERT\textsubscript{default}}) against the ones created following our method in Section \ref{method_lr_mt} (\textbf{BERT\textsubscript{prod}}). As mentioned before, we had to create 7 different LMs. Here BERT\textsubscript{prod} refers to the average performance recorded for all these LMs. For both BERT\textsubscript{default} and BERT\textsubscript{prod}, classification is achieved through stacking a linear layer on top of the corresponding language model. In the case of BERT-based LMs, the output of the first token from the final hidden state of the model is used in the final classification. Same as SVM and FT.MWPD, product metadata are concatenated into a single piece of text. The implementation and specifications of are as follows:

\begin{itemize}
	\item Implemented based on PyTorch 1.7.0\footnote{\url{https://pytorch.org/}}, with a batch size of 32, learning rate of 2e-5, epochs of 10 and using the Adam algorithm with weight decay for optimisation. All other hyperparameters remain unchanged from the distribution.
	\item For BERT\textsubscript{default}, the `bert-base-uncased' model from the generic distribution is used.
\end{itemize}


\subsubsection{Using machine translation models}\label{exp_prodcls_models_mt}
Shown in Figure \ref{fig_prodcls_structure}c (Part (c) Experiments), the baseline and their corresponding comparative models differ in terms of the product metadata used (shaded in grey). As discussed before, we apply the MT model trained in Section \ref{method_lr_mt} to product names from each dataset to generate product related keywords (\textbf{pk}), and use these keywords as another type of metadata for each product. 

For each model described above in Sections \ref{exp_prodcls_models_wordemb} and \ref{exp_prodcls_models_lm}, we first replace the product metadata with only product names, then create two variants that are compared against each other: one that uses the product name only, the other using the name (\textbf{n}) plus the product related keywords (\textbf{n,pk}, in this case the fixed length for input text is set to double that of name, i.e., 64). Also, each model only uses the generic language resources (i.e., the generic word embeddings, or the BERT LM). This is to exclude the effects from all other factors, thus allowing the results to focus on the use of product related keywords. 

As examples, SVM\textsubscript{n} is compared against SVM\textsubscript{n,pk}, both using the generic Google News word embeddings; while BERT\textsubscript{n} is compared against BERT\textsubscript{n,pk}, both using the bert-base-uncased generic LM.

\subsection{Evaluation metrics}\label{exp_evaluation_prodcls}

In terms of evaluation metrics, we use the standard Precision (P), Recall (R) and F1 scores for classification tasks. These are calculated using Equations \ref{eq1} to \ref{eq3}, where TP denotes True Positives, FP denotes False Positives, and FN denotes False Negatives. As some of our datasets contain highly unbalanced classes (e.g., MWPD-PC), we report macro-averages across all classes (arithmetic mean of individual classes' P, R, F1 scores) in order to analyse a classifier's performance on small classes, as well as weighted macro-averages (similar to macro-averages but weighs the score of each class label by the number of true instances when calculating the average) which was used in \cite{zhang2020} for ranking all participating systems. 

\begin{equation}\label{eq1}
	Precision = \frac{TP}{TP+FP}
\end{equation}

\begin{equation}\label{eq2}
	Recall = \frac{TP}{TP+FN}
\end{equation}

\begin{equation}\label{eq3}
	F1 = \frac{2 * Precision * Recall)}{Precision + Recall}
\end{equation}

\subsection{Result summary}\label{exp_prodcls_results}
In terms of the effects of using word embedding models, Tables \ref{tab_res_pc_we1} and \ref{tab_res_pc_we2} show that our skip-gram word embedding model trained on the product description corpus is able to bring consistent improvement on all datasets, with all classifiers. This improvement is noticed for Precision, Recall, and F1 (macro- and weighted macro-average), with only a handful of exceptions where the results were very close to the baseline. For example, on the Rakuten dataset, the GND.HAN\textsubscript{skip-all} obtains a macro-F1 of 37.7, which is lower but still comparable to the corresponding baseline GND.HAN\textsubscript{ggl}'s 37.8. The improvement can be significant in many cases, such as 9.0 in macro-F1 by GND.biLSTM\textsubscript{prod} against GND.biLSTM\textsubscript{ggl} on the MWPD-PC (lvl1) dataset (row 5 Table \ref{tab_res_pc_we1}), and 6.9 in macro-F1 by FT.MWPD\textsubscript{prod} against FT.MWPD\textsubscript{ggl} on the MWPD-PC (lvl2) dataset (row 11 Table \ref{tab_res_pc_we1}). The improvement on the IceCat is the smallest, but consistent. The baselines on this dataset already achieved very high F1. 

In terms of the effects of continued pre-training of the LM, Table \ref{tab_res_pc_lm} shows less promising results. We are unable to obtain consistent improvement on all datasets, but only on MWPD-PC and the IceCat datasets, where the improvement is very small. On both WDC-25 and the Rakuten datasets, the results actually declined (noticeable on WDC-25 but minor on Rakuten) with the fine tuned LM. One may argue that a potential reason for the better results on the MWPD-PC dataset is the possible similarity between the corpus used to create this gold standard, and the corpus used to continue pre-training the BERT LM. Both are based on the n-quad corpora released by the WDC project. However, we expect such impact to be minimal. On the one hand, we ensured that different releases were used (Nov 2017 release for the product description corpus, and a mixture of Nov 2018 and pre-2014 releases for the MWPD-PC goldstandard\footnote{For details, see \cite{zhang2020}.}). On the other hand, the releases were based on random crawls of the Web. Interestingly, BERT-based classifiers have achieved better results than SVM, GND based structures, or FT.MWPD on all datasets except MWPD-PC lvl3, which is harder due to more fine-grained classes. 

In terms of the effects of MT-based product keywords, Tables \ref{tab_res_pc_mt1} and \ref{tab_res_pc_mt2} show that they do not bring consistent benefits, regardless of datasets or classifiers. Although there are cases where such keywords bring improvements in the results, in the majority of cases, they caused classifier accuracy to decrease. The SVM classifier is the only one that benefited in most cases from such keywords on all datasets. Nevertheless, we cannot conclude such keywords as useful for product classification task.

\begin{landscape}
	\begin{subtables}
		\begin{table}[thb]
			\centering
			\renewcommand{\arraystretch}{1.5}
			
			\begin{tabular}{l|l c c c|c c c|l c c c| c c c}
				\hline  
				\multicolumn{15}{c}{\textbf{Part (a) Experiments: Using word embedding models}} \\	    
				\hline         
				& \multicolumn{7}{c|}{\textbf{Baselines}} &  \multicolumn{7}{c}{\textbf{Comparators}} \\	
				& & \multicolumn{3}{c}{\textbf{Macro}} & \multicolumn{3}{c|}{\textbf{W.Macro}} & &\multicolumn{3}{c}{\textbf{Macro}} & \multicolumn{3}{c}{\textbf{W.Macro}} \\	
				\hline
				Dataset   & & P   &R    &F1 &P &R	&F1 & & P   &R    &F1 &P &R	&F1    \\
				\hline
				\multirow{4}{*}{\shortstack{MWPD-PC \\ (lvl1)}}&SVM\textsubscript{ggl} &61.2    &61.5    &60.5 &80.4 &80.3	&80.2 & SVM\textsubscript{prod}& \textbf{68.6}   &\textbf{66.6}    &\textbf{66.5} &\textbf{83.0} &\textbf{82.8}	&\textbf{82.7}    \\
				&GND.biLSTM\textsubscript{ggl} &60.6    &54.5    &55.9 &81.3 &81.4	&80.8 &GND.biLSTM\textsubscript{prod}&\textbf{68.5}   &\textbf{63.2}    &\textbf{64.9} &\textbf{85.5} &\textbf{85.6}	&\textbf{85.3}    \\
				&GND.HAN\textsubscript{ggl} &66.4    &63.0    &63.0 &84.6 &83.7	&83.5 &GND.HAN\textsubscript{prod}&\textbf{71.2}   &\textbf{70.2}    &\textbf{69.9} &\textbf{86.5} &\textbf{85.9}	&\textbf{86.0}    \\
				&FT.MWPD\textsubscript{ggl} &73.3    &68.3    &69.0 &86.9 &86.5	& 86.4 &FT.MWPD\textsubscript{prod}&\textbf{76.9}   &\textbf{70.4}    &\textbf{72.3} &\textbf{87.2} &\textbf{87.5}	&\textbf{87.1}    \\
				\hline 		
				\multirow{4}{*}{\shortstack{MWPD-PC \\ (lvl2)}}&SVM\textsubscript{ggl} & 58.2   &57.8    &56.0 &80.7 &80.3	&80.2 & SVM\textsubscript{prod}& \textbf{62.4}   &\textbf{60.2}    &\textbf{58.2} &\textbf{82.6}&\textbf{82.2}	&\textbf{82.0}    \\
				&GND.biLSTM\textsubscript{ggl} &53.7    &49.9    &49.4 &80.6 &80.6	&80.1 &GND.biLSTM\textsubscript{prod}&\textbf{60.1}   &\textbf{55.2}    &\textbf{56.5} &\textbf{85.3} &\textbf{85.4}	&\textbf{85.0}    \\
				&GND.HAN\textsubscript{ggl} &57.2    &55.0    &53.5 &82.5 &82.2	&81.6 &GND.HAN\textsubscript{prod}&\textbf{61.6}   &\textbf{56.7}    &\textbf{57.0} &\textbf{86.4} &\textbf{85.9}	&\textbf{85.5}    \\
				&FT.MWPD\textsubscript{ggl} &59.5    &56.0    &55.4 &85.7 &85.0	&84.8 &FT.MWPD\textsubscript{prod}&\textbf{69.4}   &\textbf{60.9}    &\textbf{62.3} &\textbf{86.9} &\textbf{86.8}	&\textbf{86.4}    \\
				\hline 		
				\multirow{4}{*}{\shortstack{MWPD-PC \\ (lvl3)}}&SVM\textsubscript{ggl} & 47.4   &47.1    &45.2 &74.1 &73.6	&72.8 & SVM\textsubscript{prod}& 47.0   &\textbf{47.6}    &\textbf{47.0} &\textbf{77.0} &\textbf{75.8}	&\textbf{75.3}    \\
				&GND.biLSTM\textsubscript{ggl} &37.9    &37.7    &35.5 &72.0 &72.7	&71.3 &GND.biLSTM\textsubscript{prod}&\textbf{44.1}   &\textbf{43.5}    &\textbf{41.6} &\textbf{77.4} &\textbf{77.1}	&\textbf{76.2}    \\
				&GND.HAN\textsubscript{ggl} &41.9    &41.8    &39.0 &76.7 &75.6	&74.9 &GND.HAN\textsubscript{prod}&\textbf{47.5}   &\textbf{48.9}    &\textbf{45.7} &\textbf{79.1} &\textbf{78.1}	&\textbf{77.4}    \\
				&FT.MWPD\textsubscript{ggl} &41.2    &40.1    &38.5 &78.6 &75.8	&76.1 &FT.MWPD\textsubscript{prod}&\textbf{49.4}   &\textbf{46.8}    &\textbf{45.7} &\textbf{80.1} &\textbf{79.0}	&\textbf{78.5}    \\
				\hline 					
				
			\end{tabular}
			\caption{Product classification results comparing the use of word embedding models (MWPD-PC dataset). \textbf{Boldfaced} text suggests the results are better than the baseline when the language resources are used. }
			\label{tab_res_pc_we1}
		\end{table}
		
		\begin{table}[thb]
			\centering
			\renewcommand{\arraystretch}{1.5}
			
			\begin{tabular}{l|l c c c|c c c|l c c c| c c c}
				\hline  
				\multicolumn{15}{c}{\textbf{Part (a) Experiments: Using word embedding models (cont.)}} \\	    
				\hline         
				& \multicolumn{7}{c|}{\textbf{Baselines}} &  \multicolumn{7}{c}{\textbf{Comparators}} \\	
				& & \multicolumn{3}{c}{\textbf{Macro}} & \multicolumn{3}{c|}{\textbf{W.Macro}} & &\multicolumn{3}{c}{\textbf{Macro}} & \multicolumn{3}{c}{\textbf{W.Macro}} \\	
				\hline
				Dataset   & & P   &R    &F1 &P &R	&F1 & & P   &R    &F1 &P &R	&F1    \\
				\hline
				\multirow{4}{*}{IceCat}&SVM\textsubscript{ggl} & 92.5   &95.1    &93.7 &99.0 &98.8	&98.9 & SVM\textsubscript{prod}& \textbf{94.8}   &95.0    &\textbf{94.6} &\textbf{99.1} &\textbf{99.1}&\textbf{99.1}    \\
				&GND.biLSTM\textsubscript{ggl} &94.4    &94.3    &94.1 &99.2 &99.2	&99.2 &GND.biLSTM\textsubscript{prod}&\textbf{95.2}   &\textbf{94.4}    &\textbf{94.5} &\textbf{99.3} &\textbf{99.3}	&\textbf{99.3}    \\
				&GND.HAN\textsubscript{ggl} &95.0    &92.8    &93.5 &99.1 &99.0	&99.0 &GND.HAN\textsubscript{prod}&\textbf{95.6}   &\textbf{94.1}    &\textbf{94.4} &\textbf{99.2} &\textbf{99.2}	&\textbf{99.2}    \\
				&FT.MWPD\textsubscript{ggl} &93.3    &94.7    &93.7 &99.3 &99.27	&99.29 &FT.MWPD\textsubscript{prod}&\textbf{93.9}   &\textbf{95.1}    &\textbf{94.2} &\textbf{99.4} &\textbf{99.31}	&\textbf{99.33}    \\
				\hline 
				\multirow{4}{*}{WDC-25}&SVM\textsubscript{ggl} & 72.8   &73.5    &72.2 &81.8 &80.3	&80.2 & SVM\textsubscript{prod}& \textbf{74.3}  &\textbf{74.9}    &\textbf{73.9} &\textbf{82.7} &\textbf{82.0}	&\textbf{81.8}   \\
				&GND.biLSTM\textsubscript{ggl} &70.6    &70.2    &69.5 &79.1 &78.5	&78.1 &GND.biLSTM\textsubscript{prod}&\textbf{75.3}   &\textbf{76.0}    &\textbf{74.5} &\textbf{84.0} &\textbf{82.6}	&\textbf{82.5}    \\
				&GND.HAN\textsubscript{ggl} &69.1    &68.3    &67.7 &77.9 &76.7	&76.5 &GND.HAN\textsubscript{prod}&\textbf{74.3}   &\textbf{74.1}    &\textbf{73.1} &\textbf{83.2} &\textbf{81.2}	&\textbf{81.4}    \\
				&FT.MWPD\textsubscript{ggl} &77.3    &77.6    &76.5 &84.7 &84.2	&83.7 &FT.MWPD\textsubscript{prod}&\textbf{78.2}   &\textbf{78.6}    &\textbf{77.5} &\textbf{86.0} &\textbf{85.4}	&\textbf{85.0}    \\
				\hline 
				\multirow{2}{*}{Rakuten}&SVM\textsubscript{ggl}&22.5    &42.8    &26.5 &72.1 &59.4	&61.2 & SVM\textsubscript{prod}& \textbf{27.8}   &\textbf{44.0}    &\textbf{31.6} &\textbf{74.2} &\textbf{64.7}	&\textbf{66.7}    \\
				&GND.biLSTM\textsubscript{ggl} &39.8    &35.3    &35.7 &74.9 &75.1	&74.4 &GND.biLSTM\textsubscript{prod}&\textbf{41.1}   &\textbf{37.9}    &\textbf{37.8} &\textbf{76.9} &\textbf{76.7}	&\textbf{76.3}    \\
				&GND.HAN\textsubscript{ggl} &42.3    &37.7    &37.8 &76.2 &76.2	&75.5 &GND.HAN\textsubscript{prod}&42.0   &37.7    &37.7 &\textbf{76.5} &\textbf{76.4}	&\textbf{75.8}    \\
				&FT.MWPD\textsubscript{ggl} &38.6    &44.7    &39.9 &81.8 &80.6	&80.9 &FT.MWPD\textsubscript{prod}&\textbf{40.1}   &\textbf{45.7}    &\textbf{41.2} &81.8 &\textbf{80.9}	&\textbf{81.1}    \\
				\hline

			\end{tabular}
			\caption{Product classification results comparing the use of word embedding models (other datasets).  \textbf{Boldfaced} text suggests the results are better than the baseline when the language resources are used.}
			\label{tab_res_pc_we2}
		\end{table}
	\end{subtables}
	
	
	\begin{table}[thb]
		\centering
		\renewcommand{\arraystretch}{1.5}
		
		\begin{tabular}{l|l c c c|c c c|l c c c| c c c}
			\hline  
			\multicolumn{15}{c}{\textbf{Part (b) Experiments: Continued pre-training of language models}} \\	    
			\hline         
			& \multicolumn{7}{c|}{\textbf{Baselines}} &  \multicolumn{7}{c}{\textbf{Comparators}} \\	
			& & \multicolumn{3}{c}{\textbf{Macro}} & \multicolumn{3}{c|}{\textbf{W.Macro}} & &\multicolumn{3}{c}{\textbf{Macro}} & \multicolumn{3}{c}{\textbf{W.Macro}} \\	
			\hline
			Dataset   & & P   &R    &F1 &P &R	&F1 & & P   &R    &F1 &P &R	&F1    \\
			\hline
			\multirow{1}{*}{MWPD-PC (lvl1)}&BERT\textsubscript{default} & 71.3   &68.6    &68.7 &88.4 &88.4	&88.2 & BERT\textsubscript{prod}& \textbf{73.6}   &\textbf{71.8}    &\textbf{71.8} &\textbf{89.4} &\textbf{89.4}	&\textbf{89.2}    \\	
			\multirow{1}{*}{MWPD-PC (lvl2)}&BERT\textsubscript{default} & 56.7   &55.5    &54.4 &86.5 &87.1	&86.4 & BERT\textsubscript{prod}& \textbf{57.3}   &\textbf{58.1}    &\textbf{56.6} &\textbf{87.3} &\textbf{88.2}	&\textbf{87.4}    \\	
			\multirow{1}{*}{MWPD-PC (lvl3)}&BERT\textsubscript{default} & 28.1   &29.4    &26.9 & 71.1   &76.4    &72.4 & BERT\textsubscript{prod}& \textbf{31.0} &\textbf{32.0}	&\textbf{29.7}	 &\textbf{73.1}&\textbf{78.0}	&\textbf{74.3}    \\
			\hline 
			\multirow{1}{*}{IceCat}&BERT\textsubscript{default} & 97.0   &96.8    &96.8 &99.6 &99.5	&99.5 & BERT\textsubscript{prod}& \textbf{97.2}   &96.8    &96.8 &99.6 &\textbf{99.6}	&\textbf{99.6}    \\	
			\hline 
			\multirow{1}{*}{WDC-25}&BERT\textsubscript{default} & 80.5   &79.7    &79.1 &86.8 &85.6	&85.5 & BERT\textsubscript{prod}& 77.1   &77.8    &76.3 &84.4 &83.6	&83.2    \\	
			\hline 
			\multirow{1}{*}{Rakuten}&BERT\textsubscript{default} & 36.8   &34.8    &34.4 &82.0 &82.9	&82.2 & BERT\textsubscript{prod}& 36.5   &34.6    &34.1 &81.8 &82.8	&82.0    \\	
			\hline

		\end{tabular}
		\caption{Product classification results comparing the use of continued pre-training of the BERT language model.}
		\label{tab_res_pc_lm}
	\end{table}

	\begin{subtables}
		\begin{table}[thb]
			\centering
			\renewcommand{\arraystretch}{1.5}
			
			\begin{tabular}{l|l c c c|c c c|l c c c| c c c}
				\hline  
				\multicolumn{15}{c}{\textbf{Part (c) Experiments: Using machine translation models}} \\	    
				\hline         
				& \multicolumn{7}{c|}{\textbf{Baselines}} &  \multicolumn{7}{c}{\textbf{Comparators}} \\	
				& & \multicolumn{3}{c}{\textbf{Macro}} & \multicolumn{3}{c|}{\textbf{W.Macro}} & &\multicolumn{3}{c}{\textbf{Macro}} & \multicolumn{3}{c}{\textbf{W.Macro}} \\	
				\hline
				Dataset   & & P   &R    &F1 &P &R	&F1 & & P   &R    &F1 &P &R	&F1    \\
				\hline
				\multirow{5}{*}{\shortstack{MWPD-PC \\ (lvl1)}}&SVM\textsubscript{n} & 54.7   &63.7    &57.9 &79.3 &77.3	&77.9 & SVM\textsubscript{n,pk}& \textbf{60.7}   &\textbf{66.7}    &\textbf{62.8} &\textbf{80.8} &\textbf{79.9}	&\textbf{80.0}    \\
				&GND.biLSTM\textsubscript{n} &59.9    &59.8    &59.1 &81.6 &81.0	&80.8 &GND.biLSTM\textsubscript{n,pk}&\textbf{64.2}   &59.8    &\textbf{60.8} &81.6 &\textbf{81.3}	&\textbf{81.0}    \\
				&GND.HAN\textsubscript{n} &66.2    &62.5    &63.1 &81.8 &81.2	&81.1 &GND.HAN\textsubscript{n,pk}&65.2   &\textbf{63.8}    &\textbf{63.8} &\textbf{82.7} &\textbf{82.7}	&\textbf{82.4}    \\
				&FT.MWPD\textsubscript{n} &73.6    &68.1    &69.5 &87.0 &86.8	&86.6 &FT.MWPD\textsubscript{n,pk}&67.5   &63.3    &64.6 &84.7 &83.9	&84.0    \\
				&BERT\textsubscript{n} &68.6    &67.4    &67.0 &86.9    &87.2    &86.9 &BERT\textsubscript{n,pk}&68.3   &64.7    &64.6 &86.2   &86.2    &85.9    \\
				\hline 
				\multirow{5}{*}{\shortstack{MWPD-PC\\ (lvl2)}}&SVM\textsubscript{n} & 57.2   &67.6    &59.6 &79.8   &77.3    &78.0 & SVM\textsubscript{n,pk}&\textbf{58.2} &66.2	&\textbf{59.8} &\textbf{80.5} &\textbf{79.2}	&\textbf{79.4}    \\ 
				&GND.biLSTM\textsubscript{n} &59.9    &55.0    &54.8 &81.1 &80.2	&80.1 &GND.biLSTM\textsubscript{n,pk}&54.1   &49.1    &49.7 &81.0 &\textbf{80.5}	&\textbf{80.2}    \\
				&GND.HAN\textsubscript{n} &61.7    &59.4    &57.6 &81.7 &80.8	&80.6 &GND.HAN\textsubscript{n,pk}&54.8   &51.8    &51.1 &80.7 &80.4	&80.0    \\
				&FT.MWPD\textsubscript{n} &67.4    &61.7    &62.4 &86.0 &85.5	&85.3 &FT.MWPD\textsubscript{n,pk}&58.7   &54.6    &53.9 &83.7 &82.5	&82.6    \\
				&BERT\textsubscript{n} &49.4    &48.7    &47.3 &84.0 &84.8	&84.0 &BERT\textsubscript{n,pk}&48.9   &47.1    &45.4 &83.3 &84.3	&83.3    \\
				\hline 
				\multirow{5}{*}{\shortstack{MWPD-PC\\ (lvl3)}}&SVM\textsubscript{n} & 50.1   &59.0    &51.9 &77.5 &73.2	&74.0 & SVM\textsubscript{n,pk}& \textbf{51.4}   &57.7    &\textbf{52.2} &77.3 &\textbf{74.3}	&\textbf{74.6}    \\
				&GND.biLSTM\textsubscript{n} &45.2    &43.9    &42.5 &73.3 &73.1	&72.3 &GND.biLSTM\textsubscript{n,pk}&42.5   &40.8    &39.3 &72.7 &72.6	&71.5    \\
				&GND.HAN\textsubscript{n} &49.4    &47.9    &46.1 &77.2 &76.0	&75.3 &GND.HAN\textsubscript{n,pk}&48.0   &46.0    &44.9 &74.8 &75.0	&74.0    \\
				&FT.MWPD\textsubscript{n} &54.3    &53.5    &51.6 &82.7 &80.6	&80.8 &FT.MWPD\textsubscript{n,pk}&44.5   &43.8    &41.9 &78.7 &74.9	&75.8    \\
				&BERT\textsubscript{n} &23.0    &26.8    &23.4 &66.3 &73.6	&68.6 &BERT\textsubscript{n,pk}&23.4 &20.2 &23.1	&63.9 &71.4	&66.0    \\
				\hline 		
				
			\end{tabular}
			\caption{Product classification results comparing the use of MT-generated product keywords (MWPD-PC dataset).}
			\label{tab_res_pc_mt1}
		\end{table}
		
		\begin{table}[thb]
			\centering
			\renewcommand{\arraystretch}{1.5}
			
			\begin{tabular}{l|l c c c|c c c|l c c c| c c c}
				\hline  
				\multicolumn{15}{c}{\textbf{Part (c) Experiments: Using machine translation models (cont.)}} \\	    
				\hline         
				& \multicolumn{7}{c|}{\textbf{Baselines}} &  \multicolumn{7}{c}{\textbf{Comparators}} \\	
				& & \multicolumn{3}{c}{\textbf{Macro}} & \multicolumn{3}{c|}{\textbf{W.Macro}} & &\multicolumn{3}{c}{\textbf{Macro}} & \multicolumn{3}{c}{\textbf{W.Macro}} \\	
				\hline
				Dataset   & & P   &R    &F1 &P &R	&F1 & & P   &R    &F1 &P &R	&F1    \\
				
				\hline 
				\multirow{5}{*}{IceCat}&SVM\textsubscript{n} & 79.1   &91.0    &83.4 &96.9 &96.1	&96.4 & SVM\textsubscript{n,pk}& \textbf{81.6}   &91.0    &\textbf{85.1} &\textbf{97.1} &\textbf{96.5}	&\textbf{96.7}    \\
				&GND.biLSTM\textsubscript{n} & 91.3   &88.2    &89.1 &98.1 &98.0	&98.0 &GND.biLSTM\textsubscript{n,pk}& 90.7  &\textbf{88.4}    &89.0 &\textbf{98.2} &\textbf{98.1}	&\textbf{98.1}    \\
				&GND.HAN\textsubscript{n} &90.7    &86.8    &88.0 &98.0    &97.9    &97.9 &GND.HAN\textsubscript{n,pk}&\textbf{90.9}   &\textbf{87.4}    &\textbf{88.4} &\textbf{98.1}   &97.9    &\textbf{97.92}    \\
				&FT.MWPD\textsubscript{n} &91.7    &93.0    &92.1 &98.8 &98.7	&98.8 &FT.MWPD\textsubscript{n,pk}&89.8   &91.9    &90.5 &98.7 &98.6	&98.6    \\
				&BERT\textsubscript{n} &94.1    &93.2    &93.4 &98.9 &98.9	&98.9 &BERT\textsubscript{n,pk}&94.1    &93.2    &93.4 &98.9 &98.9	&98.9    \\
				\hline 
				\multirow{5}{*}{WDC-25}&SVM\textsubscript{n} & 63.7   &65.5    &63.3 &76.3 &73.6	&74.1 & SVM\textsubscript{n,pk}& 62.5   &63.8    &61.9 &73.7 &72.4	&72.3    \\
				&GND.biLSTM\textsubscript{n} &63.5    &63.3    &62.6 &74.5 &73.3	&73.2 &GND.biLSTM\textsubscript{n,pk}&\textbf{63.8}   &60.6    &60.7 &72.2 &71.9	&70.8    \\
				&GND.HAN\textsubscript{n} &62.1    &63.0    &61.2 &73.1 &71.6	&71.3 &GND.HAN\textsubscript{n,pk}&60.1   &60.5    &58.3 &71.0 &69.4	&68.9    \\
				&FT.MWPD\textsubscript{n} &69.7    &69.1    &67.5 &79.9 &78.6	&77.8 &FT.MWPD\textsubscript{n,pk}&65.8   &65.0    &63.5 &77.9 &74.5	&75.0    \\
				&BERT\textsubscript{n} &70.6    &71.0   &69.5 &80.8 &78.6	&78.7 &BERT\textsubscript{n,pk}&\textbf{71.4}   &\textbf{72.3}    &\textbf{70.7} &\textbf{81.8} &\textbf{79.9}	&\textbf{80.0}   \\
				\hline
				\multirow{5}{*}{Rakuten}&SVM\textsubscript{n} & 22.5   &42.5    &26.5 &72.1 &59.4	&61.2 & SVM\textsubscript{n,pk}& \textbf{24.1}   &41.9    &\textbf{28.0} &\textbf{74.2} &\textbf{67.4}	&\textbf{66.7}    \\
				&GND.biLSTM\textsubscript{n} &39.8    &35.3    &35.7 &74.9 &75.1	&74.4 &GND.biLSTM\textsubscript{n,pk}&38.7   &34.8    &35.0 &74.7 &75.0	&74.4    \\
				&GND.HAN\textsubscript{n} &42.3    &37.7    &37.8 &76.2 &76.2	&75.5 &GND.HAN\textsubscript{n,pk}&39.3   &34.5    &34.9 &74.6 &74.6	&73.9    \\
				&FT.MWPD\textsubscript{n} &38.6    &44.7    &39.9 &81.8 &80.6	&80.9 &FT.MWPD\textsubscript{n,pk}&36.1   &40.8    &36.8 &80.8 &79.3	&79.8    \\
				&BERT\textsubscript{n} &36.8    &34.8    &34.4 &82.0 &82.9	&82.2 &BERT\textsubscript{n,pk}&36.3   &34.2    &33.8 &81.6 &82.6	&81.8    \\
				\hline

			\end{tabular}
			\caption{Product classification results comparing the use of MT-generated product keywords (other datasets).}
			\label{tab_res_pc_mt2}
		\end{table}
	\end{subtables}

\end{landscape}

\section{Product Linking}\label{exp_prodmatch}
In this section, we explore the usage of the different language resources created in Section \ref{method_lr} in the task of product linking. Following a similar structure to Section \ref{exp_prodcls}, we present the datasets used, configuration of models, and their evaluation results. Results will be further discussed later in Section \ref{discussion} together with other tasks.

\subsection{Datasets}\label{exp_prodmatch_data}

\begin{table}[thb]
	\centering
	\renewcommand{\arraystretch}{1.5}
	
	\begin{tabular}{>{}m{2.5cm} >{}m{1.0cm} >{}m{1.3cm} >{\arraybackslash}m{1.0cm}>{\arraybackslash}m{1.5cm}>{\arraybackslash}m{2.5cm}}
		\toprule    		    		    		
		Dataset    &Train    &Validation    &Test    &Domain    &  Metadata  \\
		\hline 
		WDC-small &7,230	&1,808 	&4,400	&camera, computer, shoes, watches & name, description, brand etc. 7 types\\
		BeerAdvo-RateBeer (S) &268	&91 &91	&beer &name, brewer, style, ABV\\
		iTunes-Amazon\textsubscript{1} (S) &321	&109	&109	&music &name, artist, album etc. 8 types\\
		Fodors-Zagats (S) &567	&190	&190	&restaurant &name, address, city etc. 6 types\\
		Amazon-Google (S) &6,874	&2,293	&2,293	&software &name, manufacturer, price\\
		Walmat-Amazon\textsubscript{1} (S) &6,144	&2,049	&2,049	&electronics &name, category, brand etc. 5 types\\
		Abt-Buy (T) &5,743	&1,916	&1,916	&product &name, description, price\\
		iTunes-Amazon\textsubscript{2} (D) &321	&109	&109	&music &same as iTunes-Amazon\textsubscript{1} but misplaced\\
		Walmat-Amazon\textsubscript{2} (D) &6,874	&2,293	&2,293	&software &same as Walmat-Amazon\textsubscript{1} but misplaced\\
		\hline                          
	\end{tabular}
	\caption{Summary of datsets for product linking. (S) - structured, (T) - textual, (D) - dirty, where appropriate.}
	\label{tab_pm_dataset}
\end{table}

As shown in Table \ref{tab_pm_dataset}, we use a total of 9 datasets from two main sources: the WDC project and the DeepMatcher project. All datasets includes pairs of product offers and a binary label indicating if the offers match or not. The WDC project released several product linking datasets by parsing and annotating samples of the CommonCrawl corpus. These are used in later studies such as \cite{Peeters2020}. We use the `small' dataset as reported in \cite{Peeters2020} for a number of reasons. First, the `small', `medium', `large' and `extra large' datasets all have the same test set. The only difference is the size of the training set, which contains a different number of instances created in a distantly supervised manner. Second, our choice is also limited by our computation resources. The small WDC dataset (\textbf{WDC-small}) contains around 13k product offer pairs of four categories: computers, cameras, watches, and shoes. Each offer has the following metadata: name, description, price, brand, specification table as a text, specification key-value pairs, and site-specific label. 

The DeepMatcher project released 13 datasets for evaluating entity linking while not all of them are related to the product domain. These were further split into three groups: `structured' where product metadata are defined as key-value pairs, with values being atomic, i.e., short and pure, and not a composition of multiple values that should appear separately; `textual' where product metadata are long textual blobs (e.g., long title, short description); and `dirty' where product metadata are structured, but the values for some attributes could be misplaced, or empty. We selected 8 datasets that are arguably product-related, containing 5 structured datasets, 1 textual and 2 dirty datasets. Each dataset is split into train, test and validation sets with a ratio of 3:1:1.

Similar to the product classification datasets, the product metadata are of various word lengths and for neural network based classification models, we need to set a fixed length for them when they are used as input texts. These are configured according to Table \ref{tab_input_length_prodmatching}. All the training, validation, and test splits are based on the original data releases. Similar to the classification task, our selection of datasets is very diverse, as shown in Table \ref{tab_pm_dataset}.

\begin{table}[thb]
	\centering
	\renewcommand{\arraystretch}{1.5}
	
	\begin{tabular}{>{}m{3.2cm} >{}m{6cm}}
		\toprule    		    		    		
		Dataset    & Product metadata and fixed length (tokens)         \\
		\hline 
		WDC-small &  all (512); n (32)\\
		BeerAdvo-RateBeer (S) & all (64); n (32) \\
		iTunes-Amazon\textsubscript{1}&  all (128); n (32)\\
		Fodors-Zagats (S) &  all (64); n (32)\\
		Amazon-Google (S) &  all (64); n (32)\\
		Walmat-Amazon\textsubscript{1} (S) &  all (64); n (32)\\
		Abt-Buy (T)&  all (128); n (32)\\
		iTunes-Amazon\textsubscript{2} (D) & all (128); n (32) \\
		Walmat-Amazon\textsubscript{2} (D) &  all (64); n (32)\\
		\hline  
		
	\end{tabular}
	\caption{Configuration of input word length for neural network based  models. `all’ refers to concatenating all product metadata detailed in Table \ref{tab_pm_dataset} as a single text input.}
	\label{tab_input_length_prodmatching}
\end{table}

\subsection{Model configurations}\label{exp_prodmatching_models}
Again, since our focus is evaluating the effect of different language resources on this task, we use `out of the box' state-of-the-art solutions to configure different models using different language resources for comparison. Specifically, we use DeepMatcher \cite{Mudgal2018} and the Natural Language Inference (NLI) model based on BERT \cite{Jiang2019}. 

DeepMatcher (\textbf{DM}) is a software package\footnote{\url{https://github.com/anhaidgroup/deepmatcher}} implementing state-of-the-art entity linking algorithms using DNNs. It splits the matching process into three modules: the attribute embedding module transforms input textual data of an entity mention into word embedding-based representations; the similarity representation module learns a representation that captures the similarity of two entity mentions using their embedding representations; and the classifier module that takes as input the similarity representations to determine if the two entity mentions should match or not. The similarity representation module has two key components: attribute summarisation that implements different DNN structures for interpreting the embedding representations of input entities; and attribute comparison that implements different measures for comparing the `summary vectors' generated by the summarisation component. In this work, we configure DeepMatcher as follows:

\begin{itemize}
	\item Similarity representation module: we use a `hybrid' attribute summariser and the `element-wise absolute difference' attribute comparator, as these were found to be the optimal settings for a wide range of scenarios
	\item Classifier module: we use the multi-layer NN, which is the only option available 
	\item Attribute embedding module: this is a factor for comparison and is detailed in the section below. 
\end{itemize}

Other hyperparameters of DM remain unchanged from the default software distribution. 

For the BERT-based NLI model (\textbf{BERT}), we simply use a state-of-the-art implementation by Keras\footnote{\url{https://keras.io/examples/nlp/semantic_similarity_with_bert/}}. This model is originally created for learning sentence entailment. Given a pair of sentences, the goal of the model is to determine if the meaning of the two sentences entail each other. The model consists of two channels, each taking one sentence as input to learn a representation vector. These vectors are then concatenated and passed to a simple linear structure for classification, which determines if the two sentences entail each other or not. We simply consider each product entity as a `sentence' and construct a textual representation that fits the model. All specification and configurations remain unchanged from the implementation above.

\begin{figure}
	\centering
	\includegraphics[width=350pt]{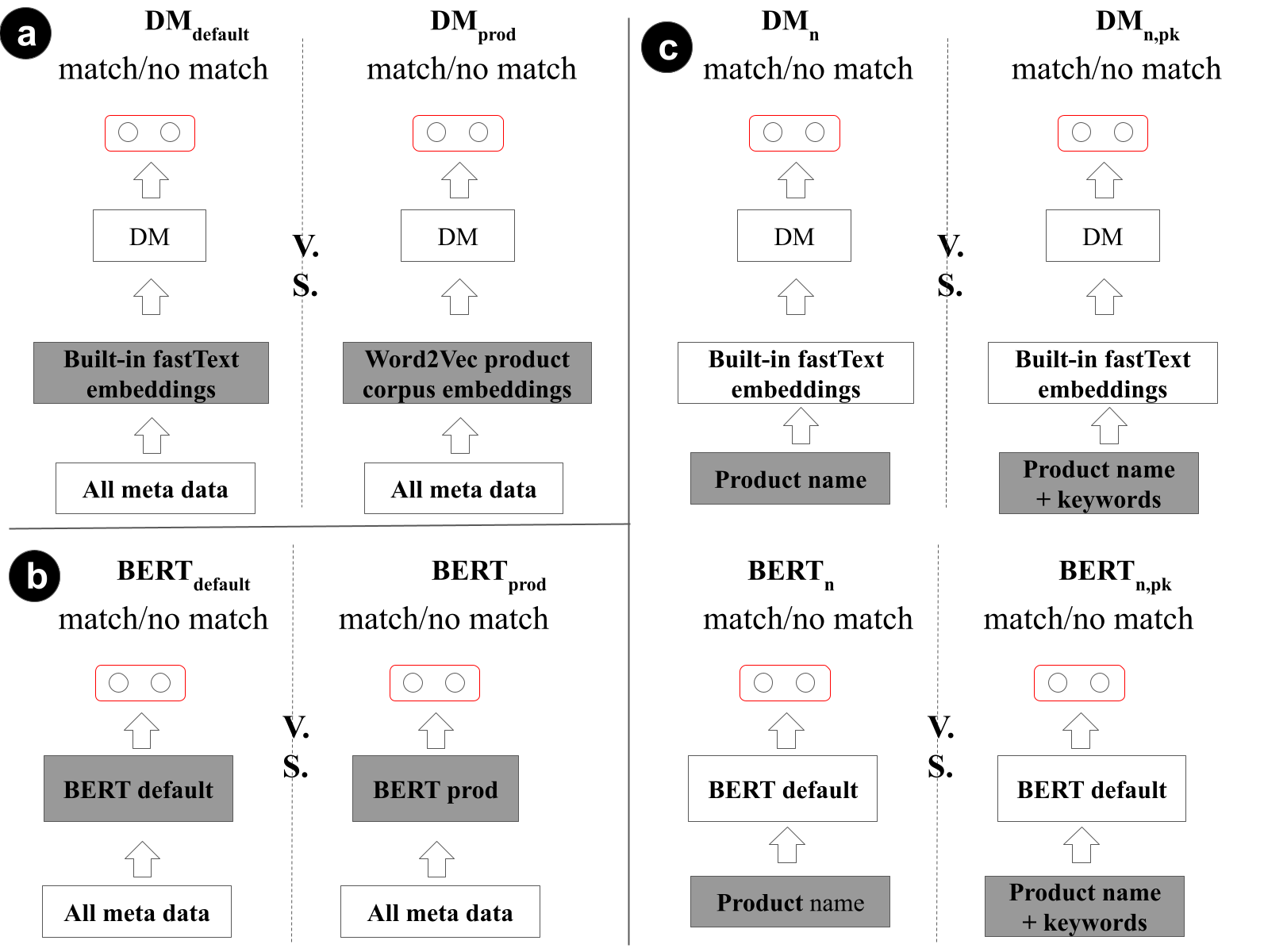}
	\caption{Configurations of different models for comparison for the product linking task. The shaded box represents the components of a model to be changed for comparison.}    
	\label{fig_prodmatching_structure} 
\end{figure}

Next, using different language resources and/or product metadata, Figure \ref{fig_prodmatching_structure} lists variants of DM and BERT to be discussed in detail below. 

\subsubsection{Using word embeddings}\label{exp_prodmatch_models_wordemb}
Shown in Figure \ref{fig_prodmatching_structure}a (Part (a) Experiments), DM using a built-in generic word embedding model (DM\textsubscript{default} baseline) is compared with DM using the product word embedding model (DM\textsubscript{prod}). Given a product, all of its metadata are concatenated into a single text input. 

\subsubsection{Using language models}\label{exp_prodmatch_models_lm}
Shown in Figure \ref{fig_prodmatching_structure}b (Part (b) Experiments), the BERT NLI model either uses the default, generic LM `bert-base-uncased' (BERT\textsubscript{default} baseline), or the product LMs (BERT\textsubscript{prod}). Same as product classification,  BERT\textsubscript{prod} refers to the average performance recorded for all the seven product LMs. Product metadata are also concatenated as a single text input. 

\subsubsection{Using machine translation models}\label{exp_prodmatch_models_mt}
Shown in Figure \ref{exp_prodmatch_models_lm}c (Part (c) Experiments), the baseline and their corresponding comparative models differ in terms of the product metadata used. Following the same way as product classification experiments, the MT model is applied to product names from each dataset to generate product related keywords (\textbf{pk}), which are used as another type of metadata for each product. Then DM and BERT (each using their generic word embeddings and LM respectively) will use either only product names as input (\textbf{n}), or product name plus product keywords as input (\textbf{n,pk}, in this case the fixed length for input text is set to 64).

\subsection{Evaluation metrics}\label{exp_prodmatching_evaluation}
Since all datasets treat the task as binary classification, we use the same evaluation metrics as product classification. The only difference is that, same as the literature, these are computed for true positives only (i.e., true matches). 

\subsection{Result summary}\label{exp_prodmatching_results}
Compared to product classification, results on product linking are even more mixed. In terms of the word embedding model (Table \ref{tab_res_pm_we}), our skip-gram based word embeddings can further improve F1 on six out of nine datasets. However, on the other three datasets they caused significant decline in F1. Referring to Table \ref{tab_pm_dataset}, we would argue that the three datasets where the decline is noted may be either too small (BeerAdvo-RateBeer (S)), or less relevant to the conventional `product' domain (Fodors-Zagats (S), restaurant; Amazon-Google (S), software). 

In terms of the continued pre-training of LM and MT-based product keywords, based on results in Tables \ref{tab_res_pm_lm} and \ref{tab_res_pm_mt}, we are unable to conclude either to be useful for this task. Positive improvements can be noticed on some datasets, but these have been very inconsistent.

\begin{table}[thb]
	\centering
	\renewcommand{\arraystretch}{1.5}
	
	\begin{tabular}{l| c c c| c c c}
		\hline  
		\multicolumn{7}{c}{\textbf{Part (a) Experiment: Using word embedding models}} \\	    
		\hline         
		& \multicolumn{3}{c|}{\textbf{Baseline (DM\textsubscript{default})}} &  \multicolumn{3}{c}{\textbf{Comparator (DM\textsubscript{prod})}} \\	
		\hline
		Dataset   &P &R	&F1  & P   &R    &F1    \\
		\hline
		\multirow{1}{*}{WDC-small} &57.3 &63.5	&60.3 &\textbf{66.2} &\textbf{65.2}	&\textbf{65.7}
		\\            		 
		\hline 
		\multirow{1}{*}{BeerAdvo-RateBeer (S)}  &66.7 &71.4	&69.0 & 57.1   &57.1    &57.1  \\            		 
		\hline 
		\multirow{1}{*}{iTunes-Amazon\textsubscript{1} (S)} &86.2 &92.6	&89.3 & \textbf{96.0}   &88.9    &\textbf{92.3}    \\            		 
		\hline 
		\multirow{1}{*}{Fodors-Zagats (S)}&100.0 &95.5	&96.7 &  90.9   &90.9    &90.9  \\            		 
		\hline 
		\multirow{1}{*}{Amazon-Google (S)} &58.8 &60.3	&59.5 &  51.0   &41.9    &46.0  \\            		 
		\hline 
		\multirow{1}{*}{Walmat-Amazon\textsubscript{1} (S)}  &33.9 &20.2	&25.3 &  \textbf{40.5}  &\textbf{23.3}    &\textbf{29.6}   \\            		 
		\hline 
		\multirow{1}{*}{Abt-Buy (T)}  &26.7 &18.9	&22.2 &  \textbf{25.5}   &\textbf{23.3}    &\textbf{24.4}    \\            		 
		\hline 
		\multirow{1}{*}{iTunes-Amazon\textsubscript{2} (D)} &57.6 &70.4	&63.3 &\textbf{72.7}   &59.3    &\textbf{65.3}   \\            		 
		\hline 
		\multirow{1}{*}{Walmat-Amazon\textsubscript{2} (D)}  &23.6 &17.1	&19.8 &\textbf{28.3}   &\textbf{24.4}    &\textbf{26.2}  \\            		 
		\hline                             
		
	\end{tabular}
	\caption{Product linking results comparing the use of word embedding models.}
	\label{tab_res_pm_we}
\end{table}


\begin{table}[thb]
	\centering
	\renewcommand{\arraystretch}{1.5}
	
	\begin{tabular}{l| c c c| c c c}
		\hline  
		\multicolumn{7}{c}{\textbf{Part (b) Experiments: Using language models}} \\	    
		\hline         
		& \multicolumn{3}{c|}{\textbf{Baseline}} &  \multicolumn{3}{c}{\textbf{Comparator}} \\	
		& \multicolumn{3}{c|}{\textbf{(BERT\textsubscript{default})}} &  \multicolumn{3}{c}{\textbf{(BERT\textsubscript{prod})}} \\	
		\hline
		Dataset   &P &R	&F1  & P   &R    &F1    \\
		\hline
		\multirow{1}{*}{WDC-small} &64.2 &83.5 &72.3 &63.2   &\textbf{85.0}    &72.1
		\\            		 
		\hline 
		\multirow{1}{*}{BeerAdvo-RateBeer (S)}  &63.3 &91.4	&73.7 & 62.5   &88.4    &71.1  \\            		 
		\hline 
		\multirow{1}{*}{iTunes-Amazon\textsubscript{1} (S)} &82.5 &86.7 &84.4 &81.6    &85.0    &82.9    \\            		 
		\hline 
		\multirow{1}{*}{Fodors-Zagats (S)}&89.0 &99.1	&93.6 &  88.7   &97.8    &92.7  \\            		 
		\hline 
		\multirow{1}{*}{Amazon-Google (S)} &68.6 &68.6	&68.0 &  64.3   &\textbf{70.5}    &66.3  \\            		 
		\hline 
		\multirow{1}{*}{Walmat-Amazon\textsubscript{1} (S)}  &66.3 &76.0	&70.5 &  58.9   &72.4    &64.4    \\            		 
		\hline 
		\multirow{1}{*}{Abt-Buy (T)}  &83.4 &69.0	&75.3 & 74.3   &\textbf{72.8}    &72.9    \\            		 
		\hline 
		\multirow{1}{*}{iTunes-Amazon\textsubscript{2} (D)} &87.0 &81.5	&84.0 &81.5   &80.4    &80.4   \\            		 
		\hline 
		\multirow{1}{*}{Walmat-Amazon\textsubscript{2} (D)}  &66.3 &69.2	&67.6 &55.2   &\textbf{71.6}    &61.2   \\            		 
		\hline                             
		
	\end{tabular}
	\caption{Product linking results comparing the use of language models.}
	\label{tab_res_pm_lm}
\end{table}


\begin{landscape}
	\begin{table}[thb]
		\centering
		\renewcommand{\arraystretch}{1.5}
		
		\begin{tabular}{l|l c c c|l c c c}
			\hline  
			\multicolumn{9}{c}{\textbf{Part (c) Experiments: Using machine translation models}} \\	    
			\hline         
			& \multicolumn{4}{c|}{\textbf{Baselines}} &  \multicolumn{4}{c}{\textbf{Comparators}} \\			
			\hline
			Dataset   & & P   &R    &F1 & &P &R	&F1     \\
			\hline
			\multirow{2}{*}{WDC-small}&DM\textsubscript{n} & 55.6   &59.9    &57.7  & DM\textsubscript{n,pk}& 51.3   &55.5    &53.3   \\   
			&BERT\textsubscript{n} &67.1    &89.2    &76.4  & BERT\textsubscript{n,pk}&66.5    &87.1    &75.3  \\            		 
			\hline 
			\multirow{2}{*}{BeerAdvo-RateBeer (S)}&DM\textsubscript{n} & 75.0   &21.4    &33.3 & DM\textsubscript{n,pk}& \textbf{83.3}   &\textbf{35.7}    &\textbf{50.0}    \\   
			&BERT\textsubscript{n} &76.2    &68.6    &68.8 & BERT\textsubscript{n,pk}&60.0    &\textbf{70.0}    &64.4   \\             		 
			\hline 
			\multirow{2}{*}{iTunes-Amazon\textsubscript{1} (S)}&DM\textsubscript{n} & 86.2   &92.6    &89.3  & DM\textsubscript{n,pk}& \textbf{100.0}   &85.2    &\textbf{92.0}    \\   
			&BERT\textsubscript{n} &86.9    &92.6    &89.5 & BERT\textsubscript{n,pk}&\textbf{87.8}  &88.9    &88.2  \\            		 
			\hline 
			\multirow{2}{*}{Fodors-Zagats (S)}&DM\textsubscript{n} & 80.0   &72.7    &76.2 & DM\textsubscript{n,pk}& 76.2   &72.7    &74.4 \\   
			&BERT\textsubscript{n} &91.3  &88.2    &89.5  & BERT\textsubscript{n,pk}&87.6    &\textbf{90.0}    &88.8  \\           		 
			\hline 
			\multirow{2}{*}{Amazon-Google (S)}&DM\textsubscript{n} & 59.5   &57.7   &58.6 & DM\textsubscript{n,pk}& 58.5   &54.7    &56.5  \\   
			&BERT\textsubscript{n} &62.5    &69.3    &64.9 & BERT\textsubscript{n,pk}&57.7    &64.8   &60.6  \\        		 
			\hline 
			\multirow{2}{*}{Walmat-Amazon\textsubscript{1}}&DM\textsubscript{n} & 36.6   &36.8    &36.7& DM\textsubscript{n,pk}& \textbf{37.2}   &18.1    &24.4   \\   
			&BERT\textsubscript{n} &61.5    &55.2    &56.5  & BERT\textsubscript{n,pk}&\textbf{63.0}    &49.7    &54.8  \\           		 
			\hline 
			\multirow{2}{*}{Abt-Buy (T)}&DM\textsubscript{n} & 33.3   &27.2    &30.0 & DM\textsubscript{n,pk}& 33.2   &\textbf{30.1}    &\textbf{31.2}   \\   
			&BERT\textsubscript{n} &76.3  &79.8    &77.9 & BERT\textsubscript{n,pk}&\textbf{81.9}   &75.6    &\textbf{78.4}    \\           		 
			\hline 
			\multirow{2}{*}{iTunes-Amazon\textsubscript{2} (D)}&DM\textsubscript{n} & 63.6   &51.9    &57.1 & DM\textsubscript{n,pk}& \textbf{70.0}   &51.9    &\textbf{59.6}    \\   
			&BERT\textsubscript{n} &81.9   &69.6   &74.3 & BERT\textsubscript{n,pk}&\textbf{87.7}   &\textbf{74.8}   &\textbf{80.4} \\    		 
			\hline 
			\multirow{2}{*}{Walmat-Amazon\textsubscript{2}}&DM\textsubscript{n} & 34.7   &34.7    &34.7& DM\textsubscript{n,pk}& 28.2   &24.9    &26.5    \\   
			&BERT\textsubscript{n} &62.5    &55.4    &56.8 & BERT\textsubscript{n,pk}&\textbf{64.6}   &51.6    &56.1    \\          		 
			\hline                             
			
		\end{tabular}
		\caption{Product linking results comparing the use of machine translation models for generating product keywords.}
		\label{tab_res_pm_mt}
	\end{table}
\end{landscape}

\section{Fake Product Review Detection}\label{exp_fakerev}
While the previous two tasks concern data that are typically properties of products, fake product review detection concerns data that is rather indirectly connected to products. For this task, we only experiment with the use of word embedding models and in-domain LMs, not the Mt model. This is because the typical review datasets do not contain product names which we require as input to the MT model. Using the review text as input will not make sense because the MT model is trained to learn mappings between short sequences of words. Also the vocabularies used during training are very different. 

\subsection{Datasets}\label{exp_fakerev_data}
We use the dataset from \cite{Salminen2022}, which were created using a Natural Language Generation model and contains over 40,000 reviews of products from 10 broad categories (e.g., Books, Electronics), each labelled as either a fake review, or genuine. For neural network based classifiers that require a fixed input text length, this is set to 512. Although the reviews are automatically generated by algorithms, the authors showed that these proved to be harder for human annotators to differentiate. We do not expand our experiments to other fake review datasets, because as we shall show later, we have observed same patterns as those in the other two tasks and we do not expect adding more datasets to add further values to our findings.

\subsection{Model configurations and evaluation metrics}\label{exp_fakerev_models_and_eval}
\begin{figure}
	\centering
	\includegraphics[width=350pt]{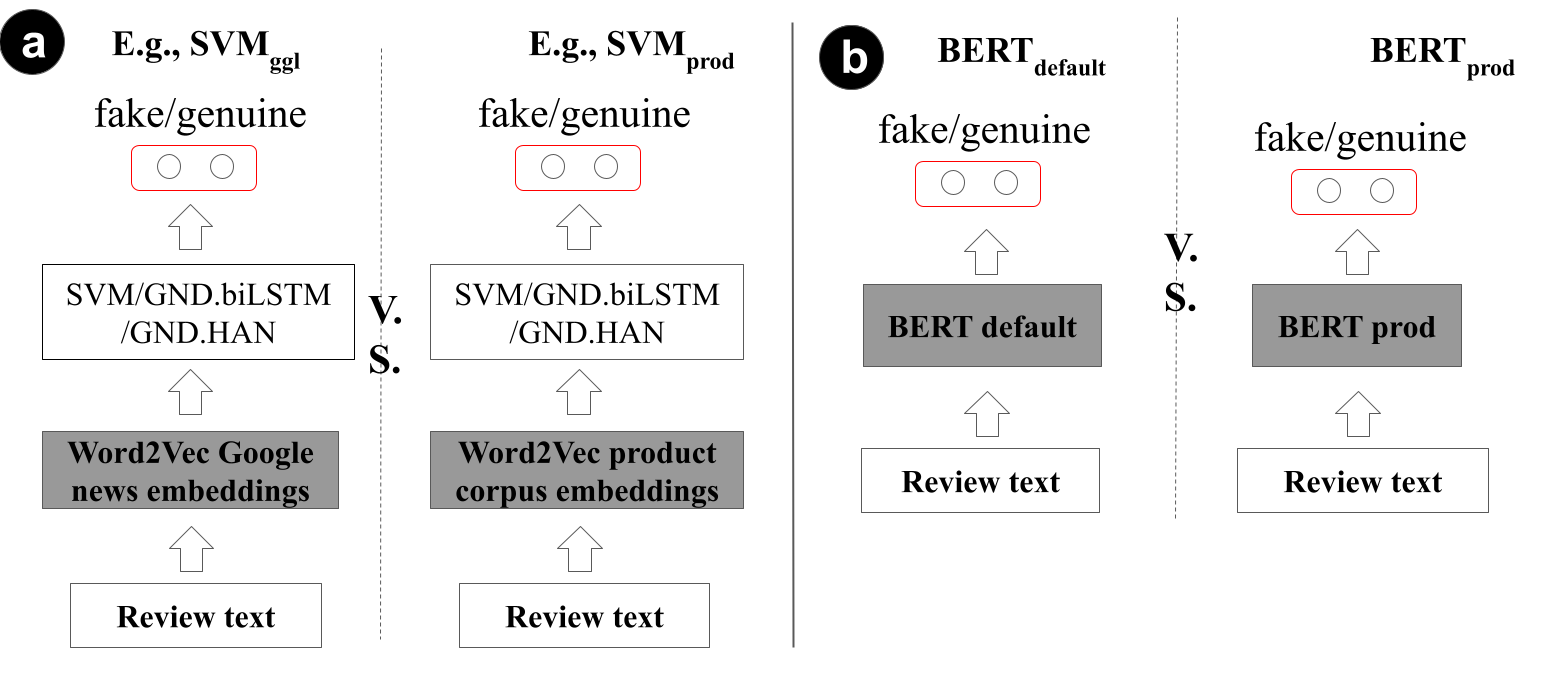}
	\caption{Configurations of different models for comparison for the fake product review detection task. The shaded box represents the components of a model to be changed for comparison.}    
	\label{fig_fakerev_structure} 
\end{figure}

Since fake review detection is a binary text classification task, our model configurations will follow those from the product classification task (Section \ref{exp_prodcls_models}). The dataset only has one source of text input, namely, the review text. Therefore, the models only vary in terms of the underlying language resources used. Figure \ref{fig_fakerev_structure} lists these models that will be briefly covered below. 

In terms of \textbf{using word embedding models}, shown in Figure \ref{fig_fakerev_structure}a (Part (a) Experiments), the baseline and their corresponding comparative models differ in terms of word embedding representations (shaded in grey). Same as product classification, we compare our product word embeddings (\textbf{prod}) against the generic, pre-trained Word2Vec embedding model (\textbf{ggl}). We combine all algorithms listed in Section \ref{exp_prodcls_models_wordemb} with each of the two options of word embedding models. The same configuration and specifications are used. 


In terms of \textbf{using in-domain LMs}, shown in Figure \ref{fig_fakerev_structure}b (Part (b) Experiments), the baseline and their corresponding comparative models differ in terms of the underlying LMs used. Again we use the same model variants from Section \ref{exp_prodcls_models_lm}, but on the different dataset. The same configuration and specifications are retained. 

In terms of \textbf{evaluation}, the same metrics for product classification explained in Section \ref{exp_evaluation_prodcls} are used here. 

\subsection{Result summary}\label{exp_fakerev_results}

\begin{table}[thb]
	\centering
	\renewcommand{\arraystretch}{1.5}
	
	\begin{tabular}{l|l c c c|c c c}
		\hline  
		\multicolumn{8}{c}{\textbf{Part (a) Experiments: Using word embedding models}} \\	    
		\hline         			
		\multicolumn{1}{c}{}&  & \multicolumn{3}{c}{\textbf{Macro}} & \multicolumn{3}{c}{\textbf{W.Macro}} \\	
		\hline
		\multicolumn{1}{c}{}   & & P   &R    &F1 &P &R	&F1    \\
		\hline
		\multirow{4}{*}{Baselines}&SVM\textsubscript{ggl} & 85.1   &85.1    &85.1 & 85.1   &85.1    &85.1     \\
		&GND.biLSTM\textsubscript{ggl} &94.8    &94.7    &94.7 &94.8    &94.7    &94.7   \\
		&GND.HAN\textsubscript{ggl} &95.0    &95.0    &95.0 &95.0 &95.0	&95.0 \\
		&FT.MWPD\textsubscript{ggl} &93.9    &93.9    &93.9 &93.9    &93.9    &93.9\\
		\hline 
		\multirow{4}{*}{Comparators}&SVM\textsubscript{prod} & \textbf{87.1}   &\textbf{87.0}    &\textbf{87.0} & \textbf{87.1}   &\textbf{87.0}    &\textbf{87.0} \\
		&GND.biLSTM\textsubscript{prod} &\textbf{95.5}    &\textbf{95.5}    &\textbf{95.4} &\textbf{95.5}    &\textbf{95.5}    &\textbf{95.4}   \\
		&GND.HAN\textsubscript{prod} &\textbf{95.8}    &\textbf{95.8}    &\textbf{95.8} &\textbf{95.8} &\textbf{95.8}	&\textbf{95.8}  \\
		&FT.MWPD\textsubscript{prod} &\textbf{94.1}    &\textbf{94.1}   &  \textbf{94.1}    &\textbf{94.1}  &  \textbf{94.1}    &\textbf{94.1}  \\
		\hline 
		\multicolumn{8}{c}{\textbf{Part (b) Experiments: Using language models}} \\	
		\hline
		Baseline&BERT\textsubscript{default}&97.3    &97.2    &97.2 &97.3    &97.2    &97.2    \\
		\hline
		Comparator&BERT\textsubscript{prod} &97.1    &97.0    &97.0 &97.1    &97.0    &97.0  \\
		\hline

	\end{tabular}
	\caption{Fake product review detection results comparing the use of word embedding models and language models.  \textbf{Boldfaced} text suggests the results are better than the baseline when the language resources are used.}
	\label{tab_res_fr}
\end{table}

Overall, we observe same patterns as product classification and linking tasks. On the one hand, product word embeddings trained on the product description corpus led to consistent improvement in F1 on this task over the generic word embedding model, with the highest noted with the SVM classifier and the lowest noted with the fastText classifer. Compared to the product classification task, it is worth highlighting that the text content in the two datasets can be notably different. This suggests that the `knowledge' captured by the product word embeddings can be potentially transferable to more general product data mining tasks. On the other hand, continued pre-training of BERT still led to detrimental effects. 

\section{Analysis and Discussion}\label{discussion}
Summarising results from Section \ref{exp_prodcls} to \ref{exp_fakerev}, we observe that building word embedding models from the product description corpus seems to be the only (arguably) effective approach to improve downstream product data mining tasks. Both continued pre-training of language models and training machine translation models could not lead to consistently better results. In this section, we reflect on these results, discuss their links to the existing literature and conduct further analysis in order to better understand our results. 

\subsection{Relation to the literature} \label{dis_lit}
In terms of our results on training word-embedding models using the product description corpus, we noted before that the earlier work by \cite{Peeters2020} used the fastText algorithm to train word embeddings using a product-related corpus collected from the semantic markup data (so called `self-trained embeddings'). They showed that this domain-specific embedding model marginally increased the performance of product linking on some product categories, but overall did not offer significant value. In comparison, our Word2Vec skip-gram word-embedding model gained notable improvement over the generic embedding model on the WDC-small dataset (Table \ref{tab_res_pm_lm}), which was also used by \cite{Peeters2020}. We also thoroughly tested it on other tasks and datasets, where we noticed consistent improvement. While authors in \cite{Peeters2020} did not study the reason of the inconsistent performance of their self-trained embeddings, the size of the corpus used for training and the different pre-processing could have attributed to this difference. The corpus used in \cite{Peeters2020} focused on product linking, and therefore, filtered the underlying data based on if a product offer contains useful product identifies. However, it is known that only a smaller percentage (less than 10\%) of product offers from the online product markup data contain this piece of information. As a result, this process could have eliminated a significant proportion of the data that may have been useful. In creating our product description corpus, we only filtered data based on the length of the text and therefore, could have retained a more diverse, and larger set of product texts. This seems to suggest that using a larger, more diverse set of product markup data is more beneficial for training word embeddings. 

In terms of our results on continued pre-training of LMs, we have obtained generally negative results: our BERT LM further pre-trained on the product description corpus underperformed the general purpose, pre-trained BERT model. While this is generally inconsistent with the wider literature on training in-domain LMs \cite{alsentzer2019,beltagy2019,Lee2019,ZhangD2020,zhang2021}, it is worth noting that similar observations were also reported in the previous literature \cite{alsentzer2019,Lee2019} - there were cases and tasks where pre-trained in-domain LMs underperformed a general purpose one. We believe there can be many reasons to this, but we speculate that the primary ones being the size and quality of the data used for in-domain pre-training. 

First, compared to the E-BERT model \cite{ZhangD2020} that is the most similar to ours, we note significant difference in terms of the pre-training process. While we used the `out of the box' BERT configuration without much change, E-BERT modified the pre-training process in many ways including: 1) providing high quality training resources in the form of well curated product domain phrases, in addition to just an unstructured corpus; 2) adding another task during the pre-training process, in addition to the standard masked language modelling task; and 3) more sophisticated masking algorithms and adaptive switching between the different masking algorithms. All these modifications allowed E-BERT to learn product related knowledge in a more effective way. Similarly, SMedBERT \cite{zhang2021} changed the pre-training process by incorporating structured information such as knowledge graphs. 

However, there exist studies that, same as ours, used a simple `out of the box' BERT pre-training process with an in-domain corpus and obtained better results on downstream tasks, such as BioBERT \cite{Lee2019}, SciBERT\cite{beltagy2019}, and  Clinical BERT \cite{alsentzer2019}. It is possible that the main differentiating factor could be the size and quality of the underlying corpora used for pre-training. All these earlier studies used resources that are arguably higher quality and are of a larger quantity. For example, SMedBERT, E-BERT and Clinical BERT used well-curated vocabularies, knowledge graphs or corpora. SciBERT and BioBERT used scientific publications that are well-written and follow a generally consistent structure. The unstructured in-domain corpus is typically of a comparable size to the original corpus used for training BERT, or much larger (BioBERT). In contrast, our corpora are much noisier as they are collected from heterogeneous websites and there are no standardisations on how the content should be written. Our corpora are also much smaller, due to having to split the entire dataset into smaller chunks to meet the computational restrictions. The computational resources used in these work were up to 8 GPUs and 23 days \cite{Lee2019}, which as acknowledged by \cite{alsentzer2019}, are still largely inaccessible to most institutions. 

One question that remains unanswered is that while both our word embedding model and the BERT LM are trained on the same corpus, why are the word embeddings more useful than the BERT model? On the one hand, the corpus used to pre-train BERT is much smaller, as we were unable to use the entire product description corpus as we did for the word embedding model. On the other hand, our word embeddings are trained using the Word2Vec skip-gram algorithm, which learns word embeddings through a task of predicting the context of a given word. The BERT LM pre-training followed the Masked Language Modelling (MLM) task, which tries to predict a word given its context. This rationale is similar to Word2Vec's Continuous Bag-of-Words (CBOW) algorithm that was shown to be less effective on modelling infrequent words \cite{Mikolov2013}. As we shall show in the following sections of data analysis and also Appendix \ref{appA}, some of the tasks we evaluated may contain many words that are under-represented in the product description corpus. As a result. This might have had an impact on the continued in-domain pre-training of BERT.

In summary, our results serve a lesson to those interested in developing large domain specific LMs following the increasing literature in this direction. While the principle seems simple, in practice, there remain many challenges. Theoretically, it appears that pre-training BERT is more susceptible to noise and size of datasets. While `more data' may be a solution to this challenge, it remains a question that `how much is enough', or to what extent this pre-training process benefits from increasing training data. At the same time, there is a lack of understanding or guidance on the `quality' aspect of the data used for training such LMs. Ultimately, how to select a `balanced' corpus, how does this matter, is there a way to focus on high quality but smaller data for pre-training LMs and if so, how should this `quality' be measured? We argue that these could be interesting questions to explore, as they may help towards addressing the inaccessibility of the current LM training approaches that are extremely computationally demanding.


Having identified the links of our results to the literature from a theoretical point view, we next develop a series of analysis to further understand the data quality issues that may contributed to our negative results. This includes a look into the data we used for training MT models in order to understand why we were unable to replicate the success of \cite{Li2018}. We present them in Sections \ref{dis_data_prov} to \ref{dis_mt_qual}.

\subsection{Data provenance}\label{dis_data_prov}
One potential cause of a less good quality training dataset is unbalanced data distribution. To understand if this could have been an issue in our study, we analysed the dominating hosts that contributed to the product description and the product category corpora in order to discover if there exist certain dominating hosts selling a restricted range of products. To do this, we manually inspected the the top 100 largest hosts as measured by the number of product offer instances found from each host, and classified them based on the types of products they sell.

\begin{table}[thb]
	\centering
	\renewcommand{\arraystretch}{1.5}
	
	\begin{tabular}{>{}m{3.2cm}>{}m{1cm} >{}m{6cm}}
		\toprule    		    		    		
		
		Type    & \#Hosts &Notes            \\
		\hline 
		auto parts/accessories &5  &E.g., car parts, car audio  \\
		automobile &7  &E.g., cars, motorcycles  \\
		B2B marketplace for exporter/importers &1  &  \\
		books &2  &Incl. conventional and audio books  \\		
		business catalogues &1  &E.g., yellow pages  \\
		camera &1  &  \\
		chemical products &1  &  \\
		educational resources &1  &E.g., posters, exercise books  \\
		farming equipment	&1	& E.g., tractors, seeds, clothing\\
		fashion	&26	&E.g., clothing, footwear, accessories\\
		finance &1	&E.g., credit card shopping\\
		food &1	& \\
		gardening	&1	&Incl. equipment, plants, decoration etc.\\
		healthy supplements & 1	&\\
		hobbies/handcrafting	&1	&E.g., knitting supplies \\
		holiday making	&4	&E.g., hotels, flights, package holidays\\
		home furnishing/furniture	&4	&\\
		information consultancy	&4	&E.g., news, drug patent consultancy, DIY\\
		jewellery/watch	&7	& \\
		lighting equipment &1	& \\
		music 	&1	&\\
		office supply	&2	&\\
		pet supply &1	&\\
		power tools &1	&\\
		e-commerce integration/comparison platforms	&5	&E.g., groupon, gumtree\\
		properties	&2	&\\
		restaurant equipment/hardware supply	&2	&\\ 
		speciality clothing &7  &All hosts are for bridal wear  \\
		sports equipment	&4	&E.g., golf, baseball\\
		sports fan shop	&2	&E.g., baseball team fan shops selling anything ranging from clothing to decoration items\\
		visual content	&4	&Typically online photo/gallery shops e.g., vectorstock.com\\
		
		\hline  
		
	\end{tabular}
	\caption{The numbers of the largest hosts (top 100 ranked by the number of product offer instances found in product description and category corpora) by types of products sold online. Note that the number will add up above 100, as there are several websites classified under multiple product types.}
	\label{tab_dis_dataprov}
\end{table}

As Table \ref{tab_dis_dataprov} shows, a significant portion of the dominating hosts sell fashion related products, typically clothing, footwear, accessories and so on. Therefore we expect that our product description and category corpora to contain a significant portion of data related to these domains. Notice that products from other domains such as software, beer, and restaurant are under-represented, which may help explain the observation that our word embeddings were not useful on the three datasets mentioned before in the product linking task. However, this analysis does not help explain why other language resources, i.e., the language model and MT-based product keywords are not as useful as word embeddings. 

\subsection{Vocabulary coverage}\label{dis_word_freq}
Here, using the product classification and linking datasets as examples, we further analyse the extent to which the vocabulary from these two tasks is represented by the product description corpus we used to train the word embeddings and the LM. We calculate a number of statistics on the training set of each dataset, and show them in Table \ref{tab_dis_wordfreq}. \textbf{Avg tok} is the average number of tokens (separated by the white space character) per instance (concatenating all metadata available) within a dataset. \textbf{Avg \% non-digit tok} is the ratio between the average number of tokens excluding tokens containing only digits and the average number of tokens per instance. \textbf{Avg \% non-digit toks in PDC} is the ratio between the average number of non-digit tokens found in the vocabulary of the product description corpus (PDC) and the average number of tokens per instance. In other words, Avg \% toks in PDC indicates how much training data are covered by the vocabulary of the product description corpus. 

\begin{table}[thb]
	\centering
	\renewcommand{\arraystretch}{1.5}
	
	\begin{tabular}{>{}m{3.5cm}>{}m{1.5cm} >{}m{3cm}>{}m{3cm}}
		\toprule    		    		    		
		
		Dataset    & Avg tok &Avg \% non-digit tok & Avg \% non-digit toks in PDC           \\
		\hline 
		MWPD-PC &80  &94\% &88\%  \\
		WDC-25 &49  &98\% &94\%  \\
		Rakuten &10  &91\% &84\%  \\
		IceCat &46  &99\% &85\%  \\
		WDC-small &12  &83\% &79\%  \\
		BeerAdvo-RateBeer (S) &17  &71\% &71\%  \\
		iTunes-Amazon\textsubscript{1} (S) &41  &63\% &59\%  \\
		Fodors-Zagats (S) &19  &42\% &43\%  \\
		Amazon-Google (S) &11  &64\% &63\%  \\
		Walmat-Amazon\textsubscript{1} (S) &17  &76\% &66\%  \\
		Abt-Buy (T) &35  &80\% &74\%  \\
		iTunes-Amazon\textsubscript{2} (D) &45  &64\% &63\%  \\
		Walmat-Amazon\textsubscript{2} (D)  &19  &74\% &69\%  \\
		\hline  
		
	\end{tabular}
	\caption{Vocabulary analysis of each dataset.}
	\label{tab_dis_wordfreq}
\end{table}

Comparing the product classification datasets against the linking datasets, there is obvious pattern that the product linking datasets contain a much larger percentage of non-alphabetic tokens (Avg \% non-digit tok), and a much smaller percentage of alphabetic tokens are covered by the product description corpus (Avg \% non-digit toks in PDC). This is likely due to the inclusion of product specification metadata such as model numbers, price, and postcode, which are much less likely to map to the vocabulary of the product description corpus. We argue that this difference could explain why our word embeddings and fine-tuned BERT language model are found to be less effective on the product linking task. 

Among the product linking dataset, recall that in Table \ref{tab_res_pm_lm} showing the results of using word embeddings, BeerAdvo-RateBeer (S), Fodors-Zagats (S) and Amazon-Google (S) are the three datasets where our word embedding model caused the baseline performance to decline. Inspecting these datasets in Table \ref{tab_dis_wordfreq}, we find them to share the following three patterns: the text content is short (Avg tok), having a large percentage of non-alphabetic tokens (e.g., 58\% for Fodors-Zagats (S), 29\% for BeerAdvo-RateBeer), and having a relatively small percentage of vocabulary coverage by the product description corpus (e.g., Avg \% non-digit toks in PDC for Fodors-Zagats (S) is only 43\%). The extreme example is the Fodors-Zagats (S) dataset. Each instance contains an average of 19 tokens, where only 8 are alphabetic words, and out of which only 3 or 4 are covered by the product description corpus. As a result, the learning algorithms could have been very sensitive to those very few number of words which are covered by the vocabulary. Adding to this the potential problem of under-representation of these domains as discussed in the previous section, the combination of these factors could explain the decline in performance when those word embeddings are used.

While other datasets may share similar patterns, they do not share all three patterns. For example, iTunes-Amazon\textsubscript{1} (S) also has a low Avg \% non-digit tok and Avg \% non-digit toks in PDC, but the text is much longer. Data from music-related websites (see Table \ref{tab_dis_dataprov}) could also have helped learning more useful word embeddings. 

\subsection{Product keywords analysis}\label{dis_mt_qual}
Here, we focus on understanding the failure of the MT-generated product keywords. We first study the quality of the MT-generated keywords. We start by showing in Table \ref{tab_dis_mtpk_example} some examples of the MT-based keywords as well as the site-specific product categories available in the MWPD-PC dataset for comparison. This shows a mixed picture of the quality of the product keywords: some could be very appropriate, while other times they can be too general, too fine-grained, or completely irrelevant.

\begin{table}[thb]
	\centering
	\renewcommand{\arraystretch}{1.5}
	
	\begin{tabular}{>{}m{3.2cm}>{}m{4cm} >{}m{2.5cm}}
		\toprule    		    		    		
		
		Name    & Site-specific product categories &MT-generated keywords            \\
		\hline 
		
		Sterling Silver Angel Charm	&All Products	&apparel accessories jewelry charms pendants\\
		HP Pavilion 23xi 58.40 cm (23ï¿½) IPS Monitor	&Product	&\\pcs workstations
		RN-XV WiFly Module - Wire Antenna 			&Home \textrangle Wireless \textrangle WiFi \textrangle RN-XV WiFly Module - Wire Antenna	&restaurant equipment ignition controls\\
		Goldwax Records T-Shirt - camel			&more sections \textrangle not available	&t shirt\\
		Thule Low Rider Fork Block 821 Pickup Truck Bike Carriers Bicycle Racks			&Vehicle Parts \& Accessories \textrangle Bike Racks \& Carriers \textrangle Bike Racks for Pickup Trucks \textrangle Thule Pick-up Truck Bike Racks	&bike racks\\
		Vintage Dining Chair			&Seating	&coastal dining chairs\\
		Mariposa Tortoise \& Stripes Sunglasses			&Accessories	&sunglasses sunglasses\\
		Helena Dress			&Clothing	&casual dresses\\
		\hline  
		
	\end{tabular}
	\caption{Examples of product names, site-specific product categories, and MT-generated product keywords (MPWD-PC dataset)}
	\label{tab_dis_mtpk_example}
\end{table}

To understand the scale of the problem, we conduct a further quantitative analysis. We look at the MWPD-PC dataset (training set only, unless otherwise stated), which is the only dataset that contains site-specific product category/labels, and previous work by \cite{Meusel2015} and \cite{Zhang2019} showed that they are useful for product classification. Thus we want to compare these site-specific product categories against those MT-generated product keywords, to see if there exists any difference that can explain why such product keywords were not as useful. Given an instance, we count the number of unique tokens (separated by white space characters) in its product name (\texttt{n}), site-specific categories (\texttt{cw}), and the MT-generated product keywords based on the name (\texttt{pk}). We then calculate two ratios: n/cw, and n/pk. Finally we plot the distribution of these two ratios for comparison. The MWPD-PC dataset contains three levels of classification, thus we repeat this on each level. 

\begin{figure}
	\centering
	\includegraphics[width=250pt]{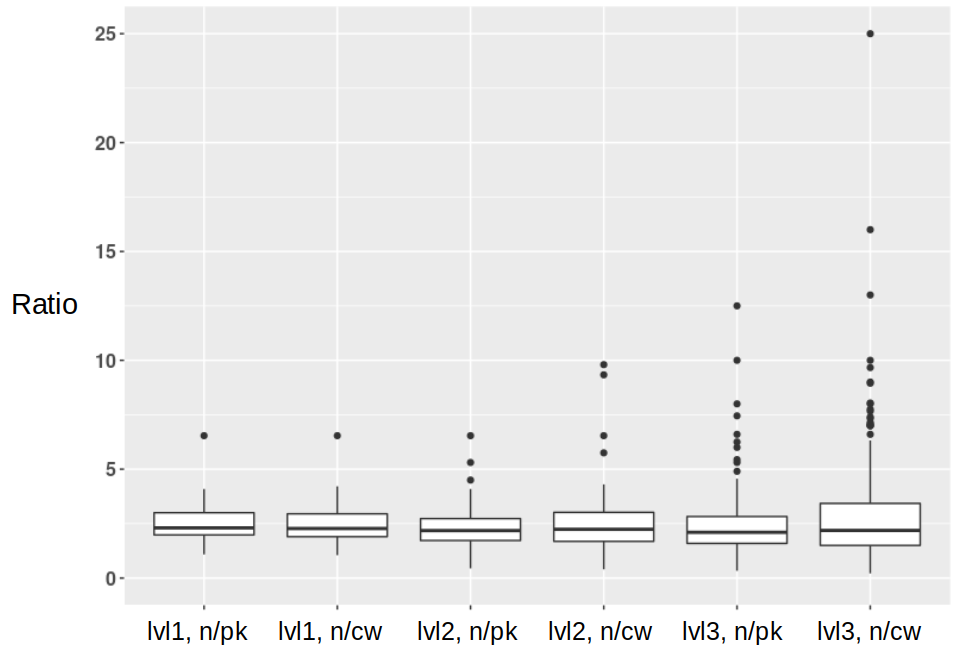}
	\caption{Comparison of the ratio between unique words from product names and that from site-specific product categories/MT-based product keywords for different levels of classification in the MWPD-PC dataset.}    
	\label{fig_2} 
\end{figure}

Figure \ref{fig_2} compares n/cw and n/pk on each level of classification in the MWPD-PC dataset. While the distributions appear to be similar on lvl1, we notice a slight difference on both lvl2 and lvl3. On both cases, n/cw have more outliers at the higher extreme compared to n/pk. Intuitively, a higher n/cw ratio suggests that a larger number of tokens in the product names relate to a smaller number of tokens in the product site-specific categories. Or in other words, it is more likely for different product instances to share a word in the site-specific product categories, which would have made it easier for the classifier to generalise. Although the difference between n/cw and n/pk is small on lvl2 and lvl3, this could help explain why the MT-based keywords were found to be marginally useful on lvl1, but not lvl2 or lvl3. 

\begin{figure}
	\centering
	\includegraphics[width=250pt]{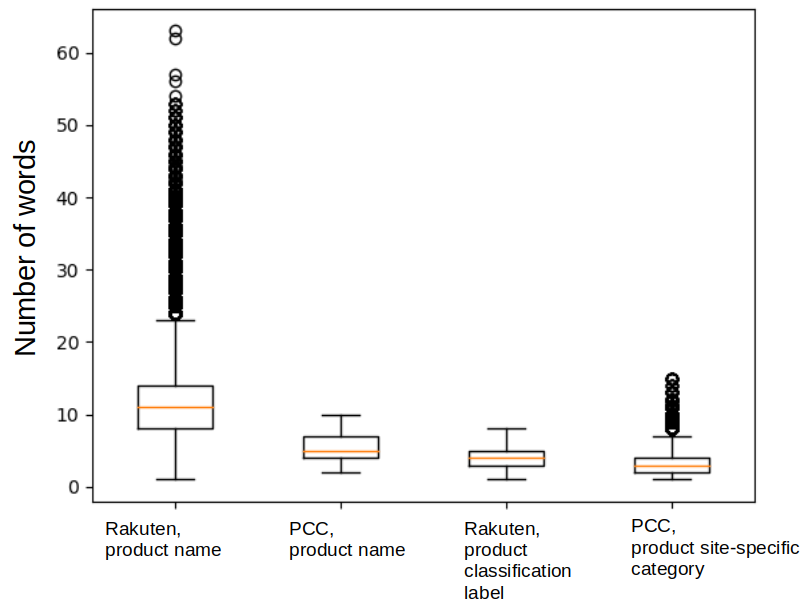}
	\caption{Comparison of the distributions of text length in the product names from the Rakuten dataset and that from the product category corpus (PCC); and comparison of the distributions of word frequency in the product classification labels from the Rakuten dataset and the word frequency in the site-specific categories of PCC.}    
	\label{fig_1} 
\end{figure}

Having understood the quality issues in the generated product keywords, we aim to understand what could have caused such issues. Our idea of MT-based product keywords is inspired by the work of \cite{Li2018}, who cast product classification as a MT task that aims to learn the mapping between product names and their category classes. They successfully tested this approach on the Rakuten dataset. Therefore we compare the Rakuten dataset against the product category corpus (PCC) to understand if there exists any difference between the two datasets that have been used to train MT models. For each instance in a dataset, we count the number of tokens in its product name, and also the number of tokens in the classification label. For the Rakuten dataset, product classifications are pseudonymised by an ID number, such as \texttt{1608 \textrangle{} 2320 \textrangle{} 2173 \textrangle{} 2878}. \cite{Li2018} treated them as a sequence of tokens. Therefore, we count the number of tokens separated by `\textrangle{}'. For the PCC, the equivalent product classifications are those site-specific product categories/labels, which we used to train the MT model. We compare the distribution of the numbers of words in Figure \ref{fig_1}. Further, we also count the number of unique tokens found in the product name and classification/site-specific categories from each dataset, and calculate a ratio as `name/category unique word ratio'. This number is 116 for Rakuten, and 4 for PCC.

Figure \ref{fig_1} shows that while the numbers of words in the product classifications from both datasets appear to be generally comparable, the Rakuten dataset however, contains generally longer product names than the PCC. The name/category unique word ratio of Rakuten is also orders of magnitude higher than that of the PCC. We say that the PCC as training data for MT is much `sparser' than the Rakuten dataset. Intuitively, it would have been easier to generalise on the Rakuten dataset, as a significantly larger number of tokens in the product names would be mapped to the same token in the product classification. On the contrary, there are significantly fewer examples to learn this mapping on the PCC. 

We conclude here that the reason behind the failure of the MT-based product keywords is the inconsistent quality of the generated keywords. And this could have been due to the sparsity in the product category corpus that we used to train the MT model.




\section{Conclusion} \label{conclusion}
In this work, we conducted an exploratory study of using structured linked data embedded within HTML webpages for the creation of language resources for downstream NLP tasks. Focusing on the e-commerce domain where such data are extremely abundant, we compared three different methods of creating such language resources: training word embedding models, continued pre-training of BERT language models, and training Machine Translation models that are used as a proxy to generate product-related keywords. These language resources are then thoroughly evaluated on three product-related NLP tasks (product classification and linking, fake product review detection) using a large number of benchmarking datasets and algorithms. Our work is the first that systematically and extensively devises, evaluates, and compares different approaches of using such structured linked data in NLP related tasks.  

Our results have shown that, surprisingly, among the three different types of language resources, only word embeddings can consistently benefit downstream tasks, while others can both damage learning accuracy of a model. While our results are generally negative, they suggest that the current popular technique of language modelling using very large unstructured corpora may not be as straightforward as the literature indicates. This raises questions such as what are the conditions on the quantity and quality of the corpora to be used for training a successful language model, how could such quality be measured, and to what extent could it compensate quantity such that the computational process is more accessible?

We further conducted analyses of the linked data we used in this study to uncover the reasons behind the mixed results, and we made several observations that may help explain these results. First, due to the highly decentralised nature in publishing such structured linked data, understandably, the published data are highly unbalanced in terms of the domains, with data related to fashion dominating a significant percentage. Second, this problem of `imbalance' could have led to noticeable vocabulary `gaps', which could be detrimental to datasets containing short texts falling into such `gaps', or algorithms that are sensitive to `infrequent' words. However, we acknowledge that these are rather `speculations', and more thorough studies may be required to investigate the true impact of these factors.

Reflecting on these findings, we discuss a few future research directions. First, our lessons learnt suggest that despite the fast-growing, significant amount of structured linked data, a `simple' approach to `use it as-is' for language resource creation may not replicate the success from earlier studies of a similar nature in other domains. While such a gigantic dataset does offer potential value, one may have to devise a careful process to select the right subset of the dataset for their tasks. This could be broadly considered an issue of quality of such linked data, but many research questions arise: how to define such quality criteria, how generic/task-specific can they be, how to use them to guide the dataset sub-selection, and what impact will it have on the language resources created using this dataset, and the downstream tasks using such resources?

Second, it may be worth to explore the use of these structured data in less domain-agnostic tasks. As an example, structured data embedded within specific HTML elements could be used as annotations on that web page, and the corresponding HTML formatting properties may be useful and more generalisable features for automatically tagging content from different web pages but formatted with similar properties. An envisaged usage scenario could be: given a collection of IMDB movie listing web pages each containing semantic markup data inserted into corresponding HTML page elements, learn the `relative' formatting features that identify the name of the movie and other metadata fields such as directors and cast members\footnote{E.g., the font size and colour of the name may be unique on the page such that it appears only once or relatively small number of times. While the metadata may appear as key-value pairs and follow a format that appears multiple times.}. Next, apply the learned model to tag individual listing pages of GoodReads, or even the more generic Amazon product listing pages. The idea here is how different types of content are formatted `relative to each other' on a product listing web page can be consistent across many different domains and websites. This has been validated in different contexts such as \cite{Potvin2019}. The abundance of already annotated product listing pages in the form of semantic markup data can create an opportunity to train such taggers in a self-supervised way. 

Our future work will explore some of the above questions and research directions.

\section*{Declarations}

\begin{itemize}
\item Funding: This work is not funded.
\item Conflict of interest: The authors declare that they have no conflict of interest.
\item Ethics approval: Not applicable 
\item Consent to participate: Not applicable 
\item Consent for publication: Not applicable 
\item Availability of data and materials: \url{https://drive.google.com/drive/folders/1BgA4iedPOFtjjAeku7c4i1c2TUu7AAHp?usp=sharing}
\item Code availability: Yes, upon request.
\item Authors' contributions: Ziqi Zhang contributed 80\% of the work (design, development, and writing); Xingyi Song contributed 20\%.
\end{itemize}

\begin{appendices}
	\section{Additional Data Analysis} \label{appA}
	
	\subsection{Product classification: word frequency analysis for word embedding model training} \label{appA_wordfreq}
	
	As discussed in Section \ref{method_lr_we}, we choose to use the Word2Vec skip-gram model instead of the continuous bag-of-words model for training the word embedding models from our product description corpus. The reason is that the skip-gram model is shown to better represent infrequent words in the training corpus. We conducted a word frequency analysis of the training datasets for product classification, and discovered that a fair percentage of words belong to the relatively infrequent segment of words in the product description corpus. We present this part of analysis here. 
	
	First, we start with extracting and normalising (same process as that used for building the word embedding model) all unique words from the product description corpus. Second, we count the frequency of these words and rank them in the descending order of frequency. This list of words is then binned into 100 segments. Third, we extract unique words from each product classification training dataset (from the concatenation of all product metadata), and count for each bin, the number of words found in that bin. Finally, we calculate the percentage of words belonging to each bin. 
	
	Our results show that, the highest frequency bin (\#1) contains roughly 71\%, 72\%, 55\% and 63\% of words found in the IceCat, Rakuten, MWPD-PC and WDC-25 training sets. In other words, between 28\% and 45\% of words from these datasets are not the most frequent words found in the word embedding training corpus. In fact, summing up the words under bin \#11 and further, these are 3\%, 6\%, 21\% and 17\% for IceCat, Rakuten, MWPD-PC and WDC-25 training sets. Based on these findings, we opted for using the skip-gram model for training word embeddings. 
\end{appendices}


\bibliography{lre2021}


\end{document}